\documentclass{article}
\PassOptionsToPackage{sort&compress,numbers}{natbib}

\usepackage[main, final]{neurips_2026}

\usepackage{amsmath,amsfonts,bm}

\usepackage{xspace}

\makeatletter
\DeclareRobustCommand{\onedot}{\futurelet\@let@token\@onedot}
\def\@onedot{%
  \ifx\@let@token.\else.\fi
  \xspace}
\makeatother

\newcommand{\eg}{\emph{e.g}\onedot}

\newcommand{\figref}[1]{Fig\onedot~\ref{#1}}

\newcommand{\secref}[1]{Sec\onedot~\ref{#1}}
\newcommand{\tabref}[1]{Tab\onedot~\ref{#1}}
\newcommand{\methodname}{Helix4D\xspace}

\def\1{\bm{1}}

\newcommand{\beas}{\begin{eqnarray*}}
\newcommand{\eeas}{\end{eqnarray*}}
\newcommand{\bea}{\begin{eqnarray}}
\newcommand{\eea}{\end{eqnarray}}

\def\mI{{\bm{I}}}

\DeclareMathAlphabet{\mathsfit}{\encodingdefault}{\sfdefault}{m}{sl}
\SetMathAlphabet{\mathsfit}{bold}{\encodingdefault}{\sfdefault}{bx}{n}

\usepackage{wrapfig}
\usepackage[utf8]{inputenc} 
\usepackage[T1]{fontenc}    
\usepackage{url}            
\usepackage{booktabs}       
\usepackage{amsfonts}       
\usepackage{nicefrac}       
\usepackage{microtype}      
\usepackage[table]{xcolor}       
\usepackage{url}
\usepackage{array}    
\usepackage{multirow} 
\usepackage{tabularx}
\usepackage{hhline}
\usepackage{booktabs}
\usepackage{graphicx}
\usepackage{xspace}
\usepackage{enumitem}
\usepackage{wrapfig}
\usepackage{floatflt}
\usepackage{pifont}
\usepackage{float}
\usepackage{graphicx} 
\usepackage{booktabs}
\usepackage{caption}
\usepackage{booktabs}
\usepackage{longtable}
\usepackage{array}
\definecolor{best}{RGB}{198,239,206}    
\definecolor{good}{RGB}{226,239,218}    
\definecolor{mid}{RGB}{255,242,204}     
\definecolor{low}{RGB}{252,229,205}     
\definecolor{worst}{RGB}{244,204,204}   
\definecolor{na}{RGB}{255,255,255}      
\definecolor{bad}{RGB}{255,210,180}
\definecolor{darkgreen}{rgb}{0.0,0.5,0.0}

\newcommand{\nav}{\cellcolor{na}{--}}

\definecolor{cvprblue}{rgb}{0.21,0.49,0.74}
\definecolor{lightcarminepink}{rgb}{0.9, 0.4, 0.38}
\usepackage[pagebackref,breaklinks,colorlinks,citecolor=cvprblue,linkcolor=lightcarminepink]{hyperref}

\newcommand{\myparagraph}[1]{\vspace*{0pt}{\noindent \bf #1}}

\title{\methodname{}: Complex 4D Mesh Generation}

\author{\textbf{Jiraphon Yenphraphai}$^{1, 2}$ \quad
        \textbf{Jianqi Chen}$^{1, 3}$\quad
        \textbf{Jian Wang}$^1$ \quad
        \textbf{Gordon Qian}$^1$ \\
        \textbf{Sergey Tulyakov}$^1$ \quad
        \textbf{Rameen Abdal}$^1$ \quad
        \textbf{Raymond A. Yeh}$^2$ \quad
        \textbf{Peter Wonka}$^{1,3}$ \quad
        \textbf{Chaoyang Wang}$^1$ \\
  $^1$Snap
  \quad
  $^2$Purdue University
  \quad
  $^3$KAUST
}

\begin{document}

\maketitle
\vspace{-1cm}
\begin{center}
{\url{https://snap-research.github.io/helix4d/}}
\end{center}

\begin{center}

    \includegraphics[width=\linewidth]{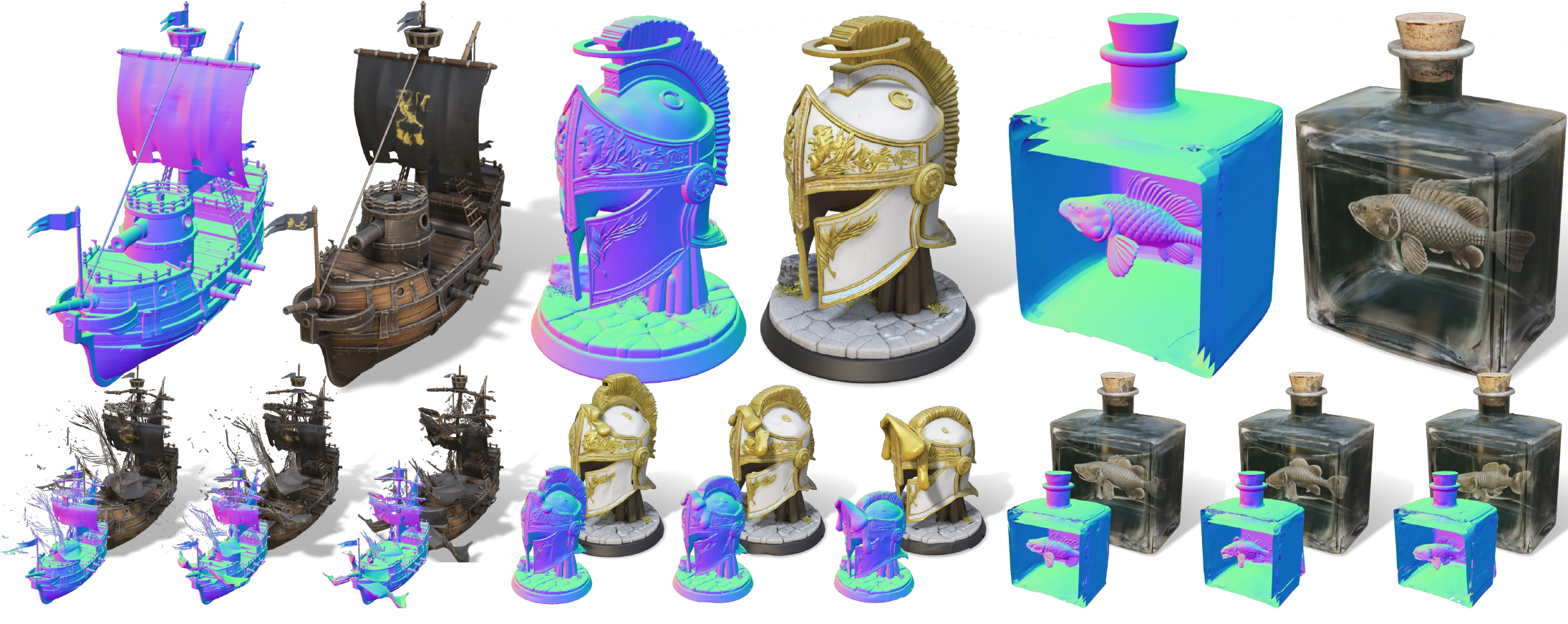}
    \captionof{figure}{
         We propose \methodname{}, a framework that can generate dynamic 3D shapes, including their materials, from an input video. \textbf{Left:} accurate reconstruction of thin structures and changes of fine details such as sails, ropes, railings, and flags breaking. \textbf{Middle:} detailed ornaments in the geometry and texture while melting. \textbf{Right:} transparent and inner-surface structures, such as a fish moving inside a glass bottle. Note, we cut the mesh open to reveal the fish inside.
         The bottom row shows generated dynamic sequences, demonstrating that \methodname{} can model challenging 4D scenarios, including topology changes, deformation, shattering, melting, transparency, and thin structures.
    }
    \label{fig:teaser}
    \vspace{-0.2cm}
\end{center}

\begin{abstract}
Current video-to-4D methods struggle with complex topology
changes, transparent materials, thin structures, and inner surfaces.  
We present \methodname{}, a dynamic mesh generation framework
by inheriting the expressive representation of Trellis2, adapting it from image-to-3D to video-conditioned 4D generation. Our design arises from two
key questions: (a) how to enable Trellis2's frame-local attention to share
information across frames while preserving its pretrained quality on rare
cases such as transparent objects and inner surfaces, and (b) how to inject
temporal information into a purely 3D positional encoding without breaking
pretrained capabilities. We address (a) with a sliding-window cross-frame
attention and anchor on the first frame. The first frame is generated by
the base Trellis2 model and injected into our model, letting it inherit
Trellis2's quality in rare cases through cross-frame attention. We address
(b) with a 4D temporal encoding that repurposes redundant
low-frequency spatial RoPE bands for time, extending the encoding from 3D with no additional parameters. Extensive experiments show the effectiveness of \methodname{} for high-quality dynamic mesh generation on ActionBench and our own challenging complex dynamics set.


\end{abstract}

\section{Introduction}
We propose a new framework for video to dynamic 3D (also called 4D) shape generation, a fundamental problem in vision and graphics, with applications in animation, virtual reality, and robotics. Compared to static 3D generation, 4D shape generation requires not only accurate geometry and materials, but also consistent modeling of motion, topology changes, and temporal coherence. 

Recent approaches to video-to-4D generation have made progress using inference-time optimization~\cite{zheng2024unified, ling2024, Yuan_2024_CVPR, bahmani2024}, multi-view video generation and reconstruction~\cite{wu2024cat4d, sun2024dimensionx, yao2025sv4d, liu2025free4d, jin2025diffuman4d}, or separate shape and motion generation~\cite{gvfd,mesh4d,motion324}.
More recently, feedforward methods that extend image-to-3D diffusion models~\cite{ss4d, shapegen4d, actionmesh} have demonstrated improved quality and generalization. However, despite strong performance on rigid and simple objects, these approaches struggle with complex topology variations, material modeling, transparency, and the reconstruction of inner surfaces.

In contrast, strong priors over geometry and materials, including the ability to represent thin structures, non-watertight surfaces, and complex appearance properties, only became available in recent large-scale foundational 3D models, \eg,  Trellis2~\cite{trellis2}.
In this work, we present a novel framework (called \methodname{}) for \emph{dynamic mesh generation from video} by systematically extending Trellis2 to 4D while preserving its pretrained strengths. Our approach enables high-quality 4D generation with significantly improved handling of challenging cases, including transparent and semi-transparent objects, complex materials, topological changes, and inner-surface reconstruction (See Fig.~\ref{fig:teaser}).

To achieve this, we address four technical challenges: enabling efficient cross-frame interaction at scale, retaining the generative ability of Trellis for challenging geometry (\eg, transparency, inner surfaces) despite limited training data, incorporating temporal information into a spatial-only positional encoding, and preserving the generative quality of the pretrained model. 
First, we design a sliding-window cross-frame attention mechanism augmented with an anchor frame and first-frame conditioning. This combines efficient local temporal interaction with a global reference signal, allowing the model to share information across frames while maintaining high-quality geometry and material quality similar to full attention. Further, conditioning on an anchor frame enables the model to retain capabilities of Trellis (transparent surfaces, inner geometry) despite very limited 4D training data of such challenging cases.
Second, we introduce a spatiotemporal rotary embedding inspired by ReRoPE~\cite{rerope}, which repurposes low-frequency spatial RoPE bands to encode time. This yields a parameter-free extension to 4D that preserves RoPE's relative-position properties and maintains compatibility with pretrained weights. A natural alternative, adopted by SS4D~\cite{ss4d}, adds a temporal RoPE~\cite{rope} on top of the backbone's existing positional encoding at every attention layer. This is suboptimal for a pretrained backbone: the extra rotation introduces phases the pretrained key/query projections never saw, disrupting the learned positional signal. We confirm this in \secref{sec:ablation}: applying the SS4D recipe in our architecture underperforms our proposed embedding.

Overall, our method converts a state-of-the-art image-to-3D generator into a video-conditioned 4D model that produces temporally consistent dynamic meshes while retaining strong geometric and material fidelity. On ActionBench~\cite{actionmesh}, \methodname{} improves CD-3D by $3.8\%$ over ActionMesh~\cite{actionmesh}. On
our harder 52-video benchmark, it outperforms all baselines on every metric,
improving ULIP-2~\cite{xue2023ulip} and Uni3D~\cite{zhou2023uni3d} by $5.7\%$ and $7.8\%$ over the strongest baseline, and is preferred over the best-performing baseline in $67.9\%$ of
user-study comparisons.
{Our main contributions are:}
\vspace{-3pt}
\begin{itemize}[topsep=0pt, leftmargin=12pt]
    \setlength{\itemsep}{0.0pt}
    \setlength{\parskip}{2.5pt}
    \item \textbf{4D generation with advanced geometric and material capabilities.} We extend a strong image-to-3D generative model to video, enabling dynamic mesh generation that for the first time handles challenging cases including transparency, complex materials, complex topology changes, and inner surface reconstruction.
    \item \textbf{Efficient cross-frame modeling with reference conditioning.} We propose a sliding-window attention mechanism with an anchor frame and first-frame conditioning. This enables efficient training and inference, as well as overcoming the data scarcity of challenging 4D training data.
    \item \textbf{Spatiotemporal RoPE via frequency repurposing.} We introduce a parameter-free extension of rotary position embeddings based on ReRoPE, which encodes time by repurposing low-frequency spatial bands while preserving relative-position properties and pretrained initialization.
 \end{itemize}

\section{Related Work}
\label{sec:related}

\myparagraph{Optimization-based 4D generation.}
Early 4D methods optimize a per-instance representation against pretrained diffusion priors. Text-conditioned variants distill motion from video diffusion via score distillation sampling~\cite{poole2022dreamfusion, singer2023text, bahmani2024, Yuan_2024_CVPR, zheng2024unified}; video-conditioned variants lift a monocular reference clip by combining photometric reconstruction with SDS, frame-interpolation, or non-rigid warping losses~\cite{jiang2023consistent4d, ren2023dreamgaussian4d, yin20234dgen, zeng2024stag4d, wang2024vidu4d, zhang20244diffusion}; two-stage methods replace distillation with multi-view video diffusion followed by 4D reconstruction~\cite{wu2024cat4d, sun2024dimensionx, yao2025sv4d, liu2025free4d, jin2025diffuman4d}; and V2M4~\cite{chen2025v2m4} registers 3D meshes into a shared topology. All four are slow (hours per asset) and have artifacts, motivating feed-forward generation.

\myparagraph{Feed-forward 3D generation.}
3D feed-forward generators differ mainly in their latent representation. Voxel-based methods such as Trellis~\citep{trellis} and Direct3D-S2~\citep{wu2025direct3d} attach features to a sparse grid intersecting the surface, while vecset-based methods originating from 3DShape2VecSet~\citep{zhang20233dshape2vecset} encode shapes as unordered latent sets decoded into an implicit field~\citep{li2025triposg, zhao2025hunyuan3d, hunyuan3d2025hunyuan3d, lai2025hunyuan3d, li2025step1x}. Both lines decode signed distance or occupancy, which requires watertight, manifold training data and cannot represent open surfaces, non-manifold topology, or inner surfaces. Trellis2~\citep{trellis2} resolves this with O-Voxels, a sparse near-surface representation that supports such structure together with PBR materials. We adopt Trellis2 as a 3D prior and adapt it to 4D.


\myparagraph{Feed-forward 4D generation.}
Existing feed-forward 4D methods split along a representation versus quality tradeoff. Deformation-centric approaches, including Mesh4D~\citep{mesh4d}, Motion 3-to-4~\citep{motion324}, ActionMesh~\citep{actionmesh}, and GVFD~\citep{gvfd}, reconstruct a canonical asset from the first frame and predict a warp field; this yields smooth motion but inherits the canonical asset's topology, which prohibits them from modeling topology changes. Other generators such as L4GM~\citep{ren2024l4gm}, SS4D~\citep{ss4d}, Sculpt4D~\cite{yin2026sculpt4d}, and ShapeGen4D~\citep{shapegen4d} learn spacetime latents end-to-end and avoid the topology constraint, but their per-frame geometry and materials qualities are limited. By building on the O-Voxel representation, \methodname{} is the first feed-forward 4D generator to support non-manifold geometry, topology changes, and transparent materials at high quality.

\section{Background}
\label{sec:prelim}

%
Trellis2~\cite{trellis2} is a 3D asset generation model that takes an input image and predicts a textured mesh. The approach comprises three main components: (i) an O-Voxel representation that converts a 3D asset into sparse voxel features, (ii) a Sparse Compression VAE  that encodes these features into a latent space, and (iii) three flow-matching models that generate sparse structure, geometry, and material latents conditioned on an input image, respectively. Given a 3D asset, Trellis2 converts it into a sparse set of active voxels: 
$
    \mathcal{F} = \{(f_i^{\mathrm{shape}}, f_i^{\mathrm{mat}}, p_i)\}_{i=1}^{L},
$
where $p_i$ is the $i^{th}$ active O-Voxels, $f_i^{\mathrm{shape}}$ stores local geometry information, and $f_i^{\mathrm{mat}}$ stores material information. Empty voxels are discarded, giving a sparse representation that is efficient for high-resolution 3D generation.
Unlike SDF-based representations~\citep{zhao2025hunyuan3d, li2025step1x, xiang2024structured}, this representation does not require watertight geometry, allowing it to represent open surfaces, thin structures, and interior surfaces. 
%
With this representation, Trellis2 generates a 3D asset from a \emph{single input image} through three flow-matching stages built on the same DiT-style backbone: a \emph{sparse-structure} stage that predicts active voxel locations, a \emph{geometry} stage that predicts per-voxel dual vertices and connectivity, and a final stage that predicts the \emph{material}. 
As this work improves the foundation of the backbone that is applied across all three stages, we describe our method generically over a single stage.

\section{Dynamic Mesh Generation}
\label{sec:method}

\begin{figure}[t]
\vspace{-0.1cm}
\small
    \centering
\includegraphics[width=\linewidth]{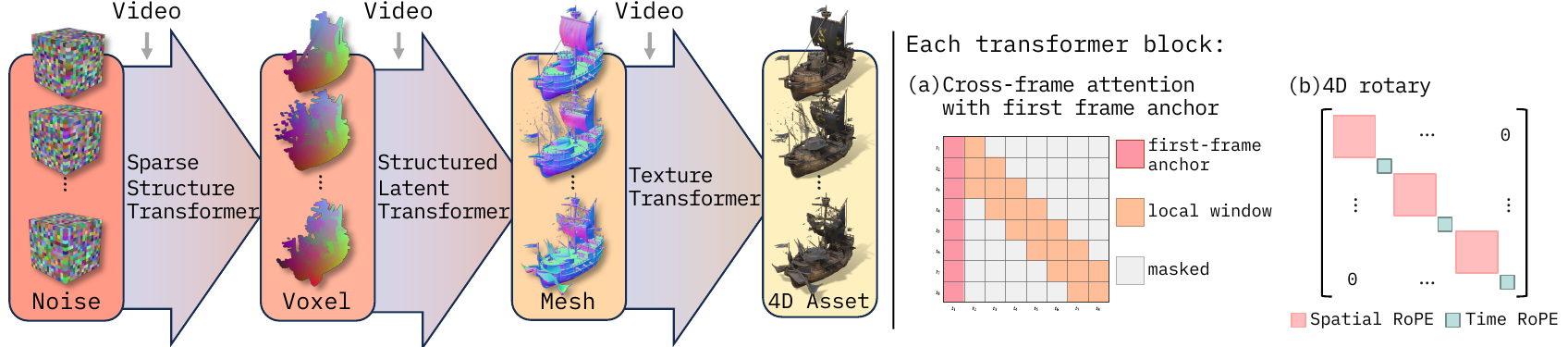}
\vspace{-0.2cm}
    \caption{\textbf{\methodname{} pipeline.} Given an input video, \methodname{} generates a 4D asset through three flow-matching stages: \emph{sparse-structure} generation, \emph{geometry} generation, and \emph{material} generation. Each self-attention layer in the transformer block uses (a) sliding-window cross-frame attention with a first-frame anchor and (b) repurposes low-frequency spatial RoPE bands for temporal encoding.}
    \label{fig:pipeline}
    \vspace{-0.4cm}
\end{figure}

The goal is to generate a dynamic textured mesh sequence from an object-centric input video. Building on the Trellis2 architecture reviewed in \secref{sec:prelim}, we generate dynamic mesh sequences from an object-centric input video by converting Trellis2 from image-to-3D into video-conditioned 4D generation while reusing its pretrained weights as in \figref{fig:pipeline}. Doing so lets us tackle cases that prior video-to-4D approaches~\citep{actionmesh, shapegen4d, mesh4d, motion324, ss4d} struggle with: complex topology changes, transparent or semi-transparent objects, and inner-surface reconstructions. \vspace{4pt}\\
{\bf Our design aims to address the following questions:} {\bf (a)}  How to enable Trellis2's frame-local attention layers to share information across frames, while preserving its pretrained generation quality on rare cases such as transparent objects and inner surfaces that are barely seen in 4D datasets? {\bf (b)} How to inject temporal information into a model whose positional encoding is purely 3D, without breaking pretrained capabilities?
We address {\bf (a)} in \secref{sec:cross_frame} with sliding-window cross-frame attention augmented by an anchor frame: the first frame is generated by the base Trellis2 model and injected into our model as the anchor, letting our model inherit Trellis2's capabilities on rare cases through cross-frame attention. We address \textbf{(b)} with a ReRoPE-inspired~\cite{rerope} temporal positional encoding in~\secref{sec:rerope}, which repurposes low-frequency spatial RoPE bands for time domains, keeping the model dimension fixed while extending the encoding from 3D to 4D.

\subsection{Cross-frame attention with first-frame anchor}

\label{sec:cross_frame}

Each Trellis2 stage operates on a sparse per-frame token sequence (reviewed in~\secref{sec:prelim}), and its pretrained attention layers are limited to only within a single per-frame sequence. To extend this to 4D, we treat the full video token stream as a \emph{sequence of sequences}, and apply cross-frame attention at every self-attention layer of the pretrained backbone.
Each frame in the input video, indexed by $f \in \{0, \dots, F-1\}$, contributes $S_f$ tokens represented by features $\{\mathbf{x}_{f,s}\}_{s=1}^{S_f}$ at voxel coordinates $\{\mathbf{p}_{f,s}\}_{s=1}^{S_f}$. The per-frame tokens are then concatenated across frames, yielding a total token sequence of length $S=\sum_{f=0}^{F-1} S_f$. Each token is identified by the pair $(f, s)$; $f=0$ denotes the first frame.

\begin{figure}[t]
\small
    \centering
\includegraphics[width=\linewidth]{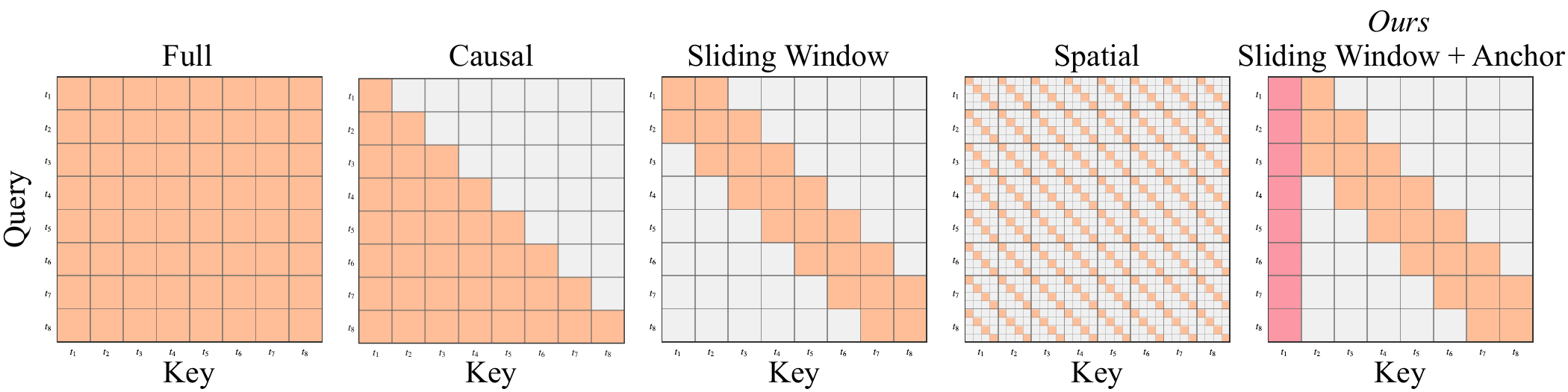}
\vspace{-0.35cm}
    \caption{\textbf{Cross-frame attention patterns.} Orange cells indicate allowed attention, gray cells indicate masked attention, and red cells indicate attention to the first-frame anchor. Full attention allows every frame to attend to each other, but it is computationally expensive. Causal attention restricts each frame to previous frames, while sliding-window attention limits attention to nearby frames for efficiency. Spatial attention attends only within corresponding spatial positions across frames. Our design combines sliding-window attention with a first-frame anchor (right), enabling efficient temporal information sharing while preserving the static 3D prior from the Trellis2 reconstruction.}
    \label{fig:attention}
    \vspace{-0.4cm}
\end{figure}

Within the transformer model, attention layers compute a query $\mathbf{q}_{f,s}$ and a key $\mathbf{k}_{f,s}$ from its feature $\mathbf{x}_{f,s}$ for each token at $(f,s)$ via learned projections. Next, how information is aggregated across tokens is based on a binary mask $\mathbf{M} \in \{0, 1\}^{S \times S}$. 
When $M_{(f,s),(f',s')} = 1$, it allows the query at $(f,s)$ to attend to the key at $(f',s')$ and $0$ blocks it. In other words, the final attention weights are computed as
\bea
    A_{(f,s),(f',s')} = M_{(f,s),(f',s')} \, \exp\!\left( \mathbf{q}_{f,s}^{\!\top} \mathbf{k}_{f',s'} / \sqrt{d} \right),
\eea
normalized over dimensions of $(f',s')$.
As illustrated in~\figref{fig:attention}, different attention designs correspond to different choices of $\mathbf{M}$. Naively using a full attention $\mathbf{M} \equiv \mI$ is too costly, as a single 4D reconstruction in our setting contains up to $10^5$ tokens. To keep the cost low, it is important to design a sparse attention pattern while allowing sufficient information to be shared across frames.


\myparagraph{Sliding window with anchor.} 
We propose to restrict the attention to a \emph{sliding window with an anchor frame}. That is, a token at frame $f$ attends to tokens within a temporal window of half-width $w$ around $f$, plus all tokens in the first frame:
\bea
    M_{(f,s),(f',s')} \;:=\;
    \begin{cases}
        1, & |f - f'| \leq w \;\text{or}\; f' = 0, \\
        0, & \text{otherwise.}
    \end{cases}
\eea
The local window captures short-range motion while the anchor (first-frame, $f^{'} = 0$) provides a shortcut for the shape context that is the most accurate in the first frame. This avoids the need to propagate shape information throughout the sequence. Empirically, sliding window with anchor attention matches full attention quality at $2\times$ lower computation cost; more discussion in~\tabref{tab:ablation_attention}.


\myparagraph{First-frame conditioning.}
\label{sec:first_frame}
A key challenge in 4D generation is data scarcity for transparent objects, semi-transparent materials, or inner surfaces' motion. A model trained from scratch on these datasets struggles to generate such properties. The pretrained Trellis2 model, by contrast, handles these objects well on static 3D assets. We feed the Trellis2-generated first frame as a clean reference, so the noisy frames $f\geq 1$ can attend to it and inherit its representations, making this task easier.

This is implemented by placing the first-frame tokens at frame index $f=0$ of the cross-frame sequence. The denoising timestep is set to zero, and frames $f\geq1$ use the standard noisy latents. During training, the flow-matching loss is computed only on frames $f\geq1$, and we use the ground-truth first-frame latent.  At sampling time, we generate the anchor by running a frozen Trellis2. 

\subsection{Repurposing Spatial RoPE for Time}
 {\bf Recap of RoPE.} Rotary Position Embedding (RoPE)~\cite{rope} encodes position into the key and query through a position-dependent rotation. Let $d$ be the feature dimension per attention head. Trellis2~\cite{trellis2} partitions these features into three equal axes for $x, y, z$, each carrying $M=d/6$ rotary frequency pairs ($d/2$ pairs total). Voxel coordinates $\mathbf{p}=(p^x, p^y, p^z)$ are integers in $[0, N-1]^3$ with grid resolution $N$, and time indices $t$ are integers from $0$ to $T-1$. Each token $m$ has an input feature $\mathbf{x}_m\in \mathbb{R}^d$ at voxel coordinate $\mathbf{p}_m$, and $\mathbf{W}_k, \mathbf{W}_q \in \mathbb{R}^{d \times d}$ denote the learned key and query projection matrices. The rotary frequencies along each axis are: 
$
    \omega_i = \theta^{-2(i-1)/M}, 
$
where $i \in \{1, \dots, M,\}$, $\theta = 10000$. We write $\mathbf{R}(\phi) = \left[\begin{smallmatrix}
\cos\phi & -\sin\phi \\
\sin\phi & \cos\phi
\end{smallmatrix}\right] \in SO(2)$
for the $2 \times 2$ rotation matrix.

\myparagraph{Spatial RoPE in Trellis2.} For a scalar position $p$ along a single axis, the per-axis rotary block is
\bea
\small
\setlength{\arraycolsep}{2pt}
    \mathbf{R}_{\Omega, p}
    = \bigoplus_{i=1}^{M} \mathbf{R}(\omega_i p)
    =
    \left[
    \begin{array}{@{}ccc@{}}
        \mathbf{R}(\omega_1 p) & & \mathbf{0} \\
        & \ddots & \\
        \mathbf{0} & & \mathbf{R}(\omega_M p)
    \end{array}
    \right]
    \in \mathbb{R}^{2M \times 2M}.
\eea
where the notation $\oplus$ means the block diagonal concatenation, and the off-diagonals are zeros.

The 3D rotary at voxel coordinate $\mathbf{p}=(p^x, p^y, p^z)$ is the block-diagonal concatenation across axes:
\bea
    \mathbf{R}_{\Omega, \mathbf{p}} \;=\; \mathbf{R}_{\Omega, p^x} \,\oplus\, \mathbf{R}_{\Omega, p^y} \,\oplus\, \mathbf{R}_{\Omega, p^z} \;\in\; \mathbb{R}^{d \times d}.
\eea
Queries and keys at token $m$ are $\mathbf{q}_m = \mathbf{R}_{\Omega, \mathbf{p}_m} \mathbf{W}_q \mathbf{x}_m$ and $\mathbf{k}_m = \mathbf{R}_{\Omega, \mathbf{p}_m} \mathbf{W}_k \mathbf{x}_m$, giving relative-position attention:
\bea
    \mathbf{q}_m^\mathsf{T} \mathbf{k}_n
    \;=\; (\mathbf{R}_{\Omega, \mathbf{p}_m} \mathbf{W}_q \mathbf{x}_m)^\mathsf{T} (\mathbf{R}_{\Omega, \mathbf{p}_n} \mathbf{W}_k \mathbf{x}_n)
    \;=\; \mathbf{x}_m^\mathsf{T} \mathbf{W}_q^\mathsf{T}\, \mathbf{R}_{\Omega,\, \mathbf{p}_n - \mathbf{p}_m}\, \mathbf{W}_k \mathbf{x}_n.
\eea
\label{sec:rerope}
\begin{figure}[t]
\small
    \centering
    \includegraphics[width=0.9\linewidth]{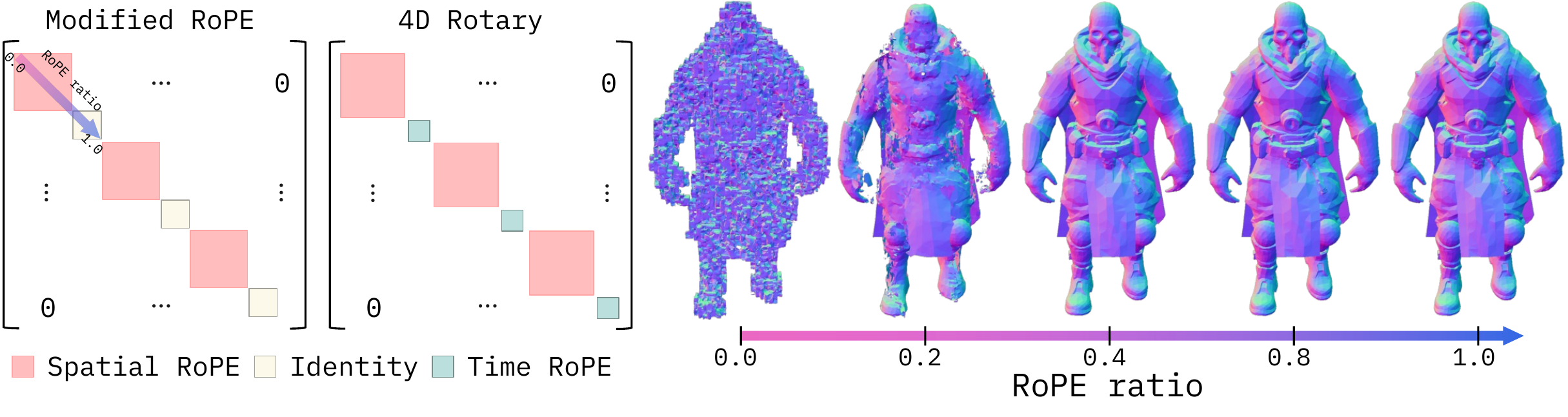}
    \vspace{-0.25cm}
    \caption{\textbf{Effect of different RoPE ratios.}
We test how many low-frequency spatial RoPE bands can be removed from a pretrained Trellis2 model by replacing them with identity matrices at inference time (Left). Too small RoPE ratios degrade geometry. In contrast, the results (Right) at ratios 0.4 and 1.0 are visually indistinguishable, suggesting that the removed low-frequency bands contribute little to the output and can be repurposed for temporal encoding without quality drop (Middle).}
    \vspace{-0.5cm}
    \label{fig:vary_rope}
\end{figure}
For 4D generation each token has a coordinate $(\mathbf{p}, t)$, where $\mathbf{p} \in \{0, \dots, N{-}1\}^3$ is the voxel coordinate and $t \in \{0, \dots, T{-}1\}$ is the frame index. We want a 4D rotary $\mathbf{R}_{\Omega, (\mathbf{p}, t)}$ that (a) reuses Trellis2's pretrained weights, (b) keeps $d$ unchanged, (c) preserves RoPE's relative-position property. A separate 1D temporal RoPE applied on top of $\mathbf{R}_{\Omega, \mathbf{p}}$ would entangle spatial and temporal phases in the same feature pairs, violating (c). Inspired by ReRoPE~\cite{rerope}, we instead repurpose existing information inside $\mathbf{R}_{\Omega, \mathbf{p}}$.

\myparagraph{Proposed re-purposing.} Our observation is that the high-frequency part of the spatial rotary matrix is sufficient to distinguish voxel coordinates, while the lower-frequency part varies slowly over the voxel grid and contributes less to spatial localization. These low-frequency channels can therefore be reused to encode time without sacrificing the generation quality. 

We test how many low-frequency bands per axis can be replaced. We split the per-axis rotary into a high-frequency block (top $\alpha M$ pairs) and a low-frequency block (bottom $(1-\alpha) M$ pairs):
\bea
    \mathbf{R}_{\Omega, p} = \mathbf{R}^{\text{high}}_{\Omega, p} \oplus \mathbf{R}^{\text{low}}_{\Omega, p} \quad \longrightarrow \quad \mathbf{R}^{\text{high}}_{\Omega, p} \oplus \mathbf{I}.
\eea
We replace the low-frequency block with identity and run inference on the Trellis2 pretrained model on a held-out 32-object validation set. As shown in \figref{fig:vary_rope}, generations is visually indistinguishable from the original ($\alpha=1$) with $\alpha \geq 0.4$, and quantitative metrics saturate across $\alpha \in [0.4, 1.0]$ (see~\tabref{tab:rerope_ratio}). When $\alpha < 0.4$, the quality drops. Therefore, any $\alpha \geq 0.4$ preserves spatial quality, and we can repurpose this low-frequency part for time embedding. Within this range, we set $\alpha = 0.75$ to allocate rotary bands proportionally to each axis's length: since the spatial extent $N=64$ and temporal extent $T=16$ are at comparable scale, the $T/N = 1/4$ ratio gives a balanced $75\%/25\%$ split between space and time.


\myparagraph{4D rotary.} Combining the truncated spatial RoPE with the phase-matched temporal block gives the full 4D rotary at $(p, t)$:
\bea
    \mathbf{R}_{\Omega, (p, t)} = \mathbf{R}^{\text{spatial}}_{\Omega, p} \oplus \mathbf{R}^{\text{temporal}}_{\Omega, t},
\eea
where $\mathbf{R}^{\text{spatial}}_{\Omega, p}$ uses the top $\alpha M$ frequency pairs along each spatial axis ($3\alpha = 3d/8$ pairs total) and $\mathbf{R}^{\text{temporal}}_{\Omega, t}$ uses the high-frequency part $N/T$-scaled frequencies on the remaining features $d/8$ pairs. 

The resulting RoPE construction gives a parameter-free extension of Trellis2's 3D RoPE to 4D space-time coordinates. Rather than adding new temporal channels or increasing the attention dimension, we allocate only the redundant low-frequency spatial bands to time. Importantly, the spatial and temporal RoPE remain separated as block-diagonal $SO(2)$ rotations; the attention map depends only on relative space-time distances.

\begin{figure}[t]
  \centering
  \vspace{-0.05cm}
  \includegraphics[width=0.98\linewidth]{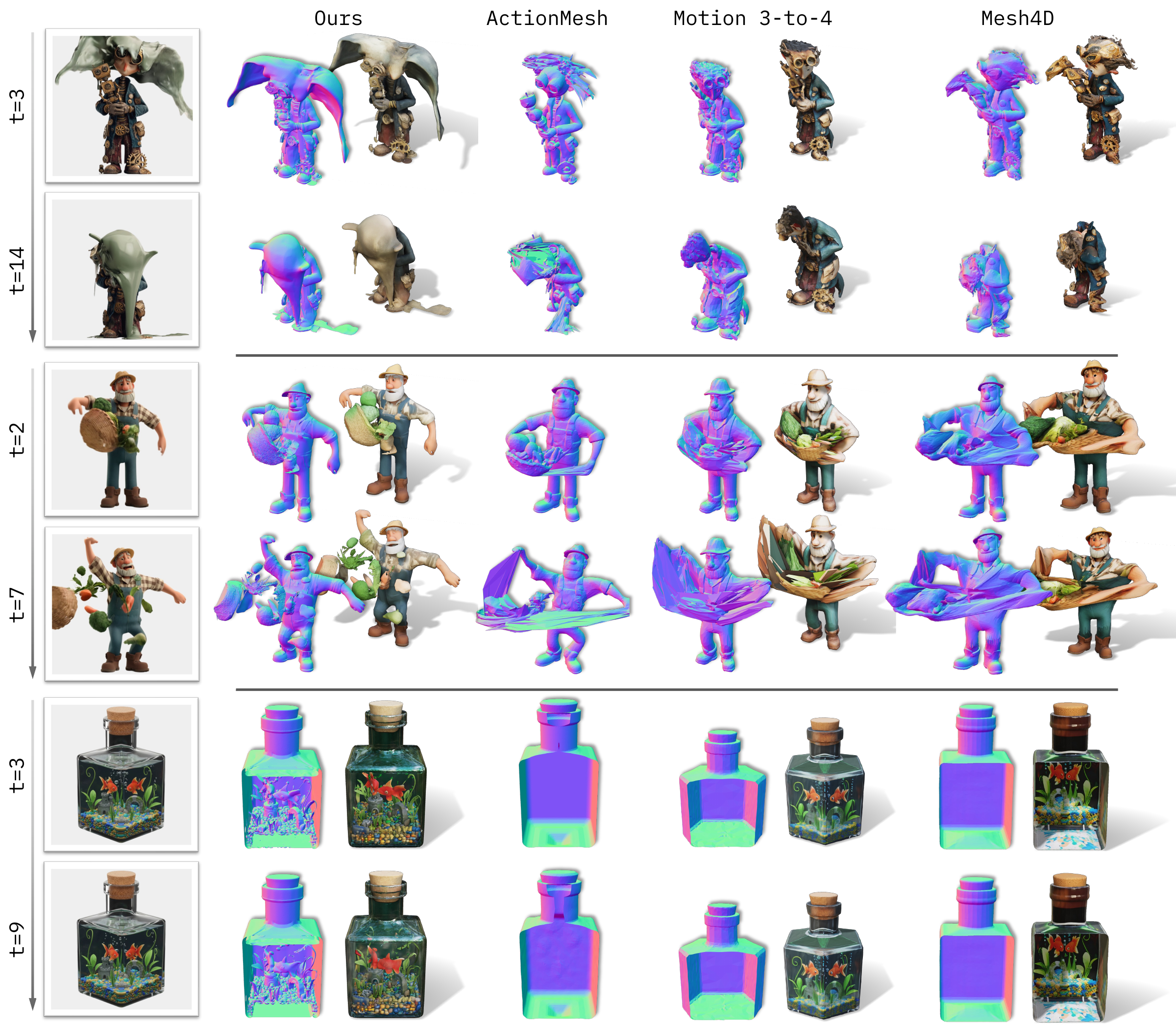}
    \vspace{-0.2cm}
  \caption{\textbf{Qualitative comparison on Helix4DBench.}
The left column shows input video frames, and each method is shown with geometry/normal renderings and, when available, shading. Compared with ActionMesh~\cite{actionmesh}, Motion 3-to-4~\cite{motion324}, and Mesh4D~\cite{mesh4d}, our method better preserves fine geometry, topology changes, emerging structures, and transparent objects.}
  \label{fig:results}
  \vspace{-0.5cm}
\end{figure}
\section{Experiments}
\label{sec:experiment}
We evaluate our method on 4D generation against five
video-to-4D baselines~\cite{ss4d, shapegen4d, mesh4d, motion324, actionmesh},
on both our newly introduced 52-video benchmark, covering topology change,
transparency, and volumetric phenomena, and
ActionBench~\cite{actionmesh}. We then ablate each of our three core
design choices and analyze alternative cross-frame attention patterns
(\secref{sec:ablation}). 

\myparagraph{Data curation.} We curate our training data from the subset of animated TexVerse-1K~\cite{texverse}, which has about 55k objects. For each asset, we extract 16 animation frames and convert every frame into an O-voxel representation at a resolution of $1024^3$, following Trellis2~\cite{trellis2}. To ensure consistent scale and position across time, each object is rescaled using the union of per-frame bounding boxes so that the entire animation lies within $[-0.5, 0.5]$. For each animation, we render 16 views from randomly sampled camera viewpoints on the sphere looking at the origin at a resolution of $1024 \times 1024$, with randomized focal length, radius, azimuth, and elevation.

\myparagraph{Architecture and training details.} 
We apply our method (4D rotary, our proposed attention, and first-frame conditioning) uniformly to all three Trellis2 stages: sparse-structure, geometry, and material. Each stage generates 16 frames jointly. 4D rotary keeps $\alpha=0.75$ of the spatial frequencies, and the rest are repurposed for time. Cross-frame attention uses a window size of $5$ plus the first frame as an anchor. Each stage fine-tunes only self-attention layers from the pretrained Trellis2 with a batch size of $32$ on $32$ A100 GPUs for $20$K iterations, 
using AdamW~\cite{loshchilov2017adamw} with learning rate $2 \times 10^{-5}$.

\subsection{Comparisons}

\myparagraph{Test set.} As no public benchmark focuses on complex
dynamic 4D generation, we construct \emph{Helix4DBench}, a 52-video test set covering morphing,
emerging objects, shattering, transparent and translucent objects, and
volumetric phenomena such as smoke and fire. We source still images from
publicly available Trellis2~\cite{trellis2} examples, and animate each into a
16-frame video using Wan2.2~\cite{wan2025}, and remove backgrounds with
rembg; full construction details are provided in the supplementary. To compare against prior video-to-4D approaches on their benchmark, we evaluate on ActionBench~\cite{actionmesh}. 

\myparagraph{Baselines.} We compare against five recent methods for dynamic 3D generation on 16-frame videos: SS4D~\cite{ss4d}, ShapeGen4D~\cite{shapegen4d}, Mesh4D~\cite{mesh4d}, Motion 3-to-4~\cite{motion324} and ActionMesh~\cite{actionmesh}. Some baselines do not output texture (ActionMesh, ShapeGen4D), and we mark texture-dependent metrics as `\nav{}' for these methods. Mesh4D$^\ast$ supports only six output frames; therefore, for a
fair comparison with our 16-frame setting, we uniformly sample six frames from each generated sequence.

\myparagraph{Evaluation metrics.}
\begin{table}[t]
  \caption{\textbf{Quantitative comparison on our Helix4DBench.} Ours outperforms prior work across all reported metrics, including 3D alignment, video quality, temporal consistency, and pairwise user preference.
Preference reports the 1-on-1 win rate of ours against each baseline.}
\label{tab:quantitative_comparison_ours}
  \small
  \centering
  \setlength{\tabcolsep}{7pt}
  \renewcommand{\arraystretch}{1.2}
  \resizebox{.9\linewidth}{!}{
  \begin{tabular}{l c c c c c c c}
    \specialrule{.15em}{.05em}{.3em}
    \multirow{2}{*}{\textbf{Method}}
    & \multicolumn{4}{c}{\textbf{Alignment}}
    & \multicolumn{2}{c}{\textbf{Video Quality}}
    & \multicolumn{1}{c}{\textbf{Preference}} \\
    \cmidrule(lr){2-5} \cmidrule(lr){6-7} \cmidrule(lr){8-8}
    & \textbf{CLIP}$\uparrow$
    & \textbf{CLIP-N}$\uparrow$
    & \textbf{ULIP-2}$\uparrow$
    & \textbf{Uni3D}$\uparrow$
    & \textbf{DreamSim}$\downarrow$
    & \textbf{FVD}$\downarrow$
    & \textbf{Ours win rate}$\uparrow$ \\
    \midrule

    SS4D~\cite{ss4d}
    & 0.8468
    & \nav
    & 0.4191
    & 0.3739
    & 0.2745
    & 1046
    & \textbf{71.2\%} \\

    ShapeGen4D~\cite{shapegen4d}
    & \nav
    & 0.6876
    & 0.4169
    & 0.3687
    & \nav
    & \nav
    & \textbf{67.9\%} \\

    Mesh4D$^\ast$~\cite{mesh4d}
    & 0.8533
    & 0.7399
    & 0.4353
    & 0.3691
    & 0.3025
    & 979
    & \textbf{79.8\%} \\

    Motion 3-to-4~\cite{motion324}
    & 0.8674
    & 0.7442
    & 0.4331
    & 0.3768
    & 0.2697
    & 955
    & \textbf{80.7\%} \\

    ActionMesh~\cite{actionmesh}
    & \nav
    & 0.7156
    & 0.4226
    & 0.3761
    & \nav
    & \nav
    & \textbf{89.8\%} \\

    \midrule
    Ours
    & \textbf{0.8723}
    & \textbf{0.7569}
    & \textbf{0.4600}
    & \textbf{0.4063}
    & \textbf{0.2593}
    & \textbf{950}
    & Ref. \\
    \specialrule{.15em}{.05em}{.05em}
  \end{tabular}}
  \vspace{-0.5cm}
\end{table}
Our Helix4DBench is constructed by animating still images with Wan2.2~\cite{wan2025} and therefore has no ground-truth geometry. We follow Trellis2~\cite{trellis2} for static-frame quality and add temporal-consistency metrics for the 4D setting. When ground-truth geometry is available (\eg, on ActionBench~\cite{actionmesh}), we report Chamfer distance following ActionMesh~\cite{actionmesh}, as introduced later in this section. CLIP~\cite{radfordclip} and CLIP-N measure appearance and geometry quality, respectively, computed as the similarity between rendered views (or normal maps) at azimuths $\{0^\circ, 90^\circ, 180^\circ, 270^\circ\}$ and the ground-truth video frames. ULIP-2~\cite{xue2023ulip} and Uni-3D~\cite{zhou2023uni3d} measure 3D-image alignment by sampling 10K points from the mesh surface via farthest-point sampling and computing similarity to the groundtruth image. Baselines without textures are assigned a uniform gray color for fairness. DreamSim~\cite{fu2023dreamsim} is a learned perceptual similarity metric that aligns with human judgments on pose, layout, color, and semantics differences. We apply it per-frame between rendered and ground-truth images and average across frames. FVD~\cite{unterthiner2019fvd} measures the temporal consistency of the rendered images. We also conduct a user study. Participants compare our result against one baseline in a pairwise A/B setting and select the better result in terms of overall visual quality, geometric detail, and temporal consistency. We report the win rate of ours against each baseline.

\myparagraph{Quantitative comparisons on Helix4DBench.} 
\tabref{tab:quantitative_comparison_ours} reports results on our benchmark. Our method outperforms all baselines across every metric, indicating both higher per-frame quality and stronger temporal consistency. These quantitative gains are consistent with the qualitative results in~\figref{fig:results}, which shows three modes that distinguish our approach from ActionMesh~\cite{actionmesh}, Motion 3-to-4~\cite{motion324}, and Mesh4D~\cite{mesh4d}: \vspace{3pt}\\
\textit{(i) Emerging objects.}
 They all treat the first frame as an anchor and build motion on top of an initial mesh. They therefore fail whenever new content appears in later frames. The first example in~\figref{fig:results} shows paint splashing onto and partially melting the figure. The content that is absent in the first frame is thus unrecoverable for anchor-based baselines.\vspace{3pt}\\
\textit{(ii) Vertex sticking.}
Even when an object is present in the first frame, anchor-based methods reconstruct it as a single mesh and propagate vertices through time. When the object needs to be separated later, this leads to a vertex-sticking problem as visible in the farmer-with-vegetable-basket example: the basket fuses with the hands and distorts as the figure moves.\vspace{3pt}\\
\textit{(iii) Inner surfaces.}
Baselines rely on a watertight-mesh assumption and therefore cannot represent inner surfaces. The fish-in-bottle example illustrates this: only our method reconstructs the transparent shell together with the fish swimming inside. 

\myparagraph{Quantitative comparisons on ActionBench and TexVerse test set.} We follow ActionMesh~\cite{actionmesh} for geometric accuracy, given a predicted mesh sequence and the ground-truth point clouds, 
\begin{wraptable}[13]{r}{0.59\textwidth}
  \centering
  \small
  \setlength{\tabcolsep}{4pt}
    \vspace{-0.4cm}
    \centering
    \caption{\textbf{Quantitative comparison on ActionBench and TexVerse test set.}
    Ours achieves the lowest CD-3D on both benchmarks, and the lowest CD-4D on the TexVerse set.}
    \label{tab:cd_comparison}
    \resizebox{\linewidth}{!}{
    \begin{tabular}{lcccc}
      \toprule
      \multirow{2}{*}{\textbf{Method}}
      & \multicolumn{2}{c}{\textbf{ActionBench}}
      & \multicolumn{2}{c}{\textbf{TexVerse Test Set}} \\
      \cmidrule(lr){2-3} \cmidrule(lr){4-5}
      & \textbf{CD-3D} $\downarrow$
      & \textbf{CD-4D} $\downarrow$
      & \textbf{CD-3D} $\downarrow$
      & \textbf{CD-4D} $\downarrow$ \\
      \midrule
      SS4D~\cite{ss4d}
        & 0.105 & 0.120
        & 0.052 & 0.103 \\
      ShapeGen4D~\cite{shapegen4d}
        & 0.056 & 0.170
        & 0.075 & 0.129 \\
      Mesh4D$^\ast$~\cite{mesh4d}
        & 0.096 & 0.141
        & 0.058 & 0.100 \\
      Motion 3-to-4~\cite{motion324}
        & 0.068 & 0.114
        & 0.056 & 0.108 \\
      ActionMesh~\cite{actionmesh}
        & 0.053 & \textbf{0.081}
        & 0.056 & 0.111 \\
      \midrule
      \textbf{Ours}
        & \textbf{0.051} & 0.093
        & \textbf{0.045} & \textbf{0.097} \\
      \bottomrule
    \end{tabular}}
\end{wraptable}%
we sample dense point clouds from the predicted meshes and align them to the ground truth via Iterative Closest Point (ICP). CD-3D computes the average per-frame Chamfer distance after per-frame ICP
alignment, measuring shape accuracy. CD-4D estimates a single ICP alignment from the first frame and applies it uniformly to every subsequent frame, measuring both shape accuracy and temporal consistency. \vspace{4pt}\\
Note, ActionBench contains mainly simple motion, non-topology-changing sequences, the settings ActionMesh is specifically designed for.
Hence, we also evaluate on a held-out subset of TexVerse-1K, which has broader dynamic scenarios. 
We randomly sample 32 objects from this held-out subset and use the ground-truth meshes for CD evaluation.
\tabref{tab:cd_comparison} reports CD-3D and CD-4D on both benchmarks. 
On ActionBench, our method achieves the lowest CD-3D, indicating the highest
per-frame geometry accuracy. 
On our harder TexVerse, our method achieves
the best results on \emph{both} CD-3D and CD-4D, confirming that the gains
generalize to sequences with topological and material
complexity that existing methods struggle to handle. Baselines exhibit vertex sticking and fail to follow the motion when topology changes (\eg, the opening door fuses shut); qualitative comparisons provided in~\figref{fig:results_complex}.

\subsection{Ablation}
\label{sec:ablation}
\begin{figure}[t]
  \centering

  \begin{minipage}[c]{0.42\linewidth}
    \centering
    \captionof{table}{\textbf{Component ablation.}
    The full model is best across all metrics.}
    \label{tab:ablation}

    \renewcommand{\arraystretch}{1.25}
    \resizebox{\linewidth}{!}{
    \begin{tabular}{lccc}
      \toprule
      \textbf{Variant}
      & \textbf{CD-3D} $\downarrow$
      & \textbf{ULIP-2} $\uparrow$
      & \textbf{Uni3D} $\uparrow$ \\
      \midrule
      w/o first-frame cond.
        & 0.0488 & 0.2809 & 0.2554 \\
      w/o 4D rotary
        & 0.0478 & 0.2819 & 0.2584 \\
      w/o our attention
        & 0.0467 & 0.2819 & 0.2530 \\
      \midrule
      Ours
        & \textbf{0.0464}
        & \textbf{0.2881}
        & \textbf{0.2589} \\
      \bottomrule
    \end{tabular}}
  \end{minipage}
  \hfill
  \begin{minipage}[c]{0.55\linewidth}
    \centering
    \includegraphics[width=\linewidth]{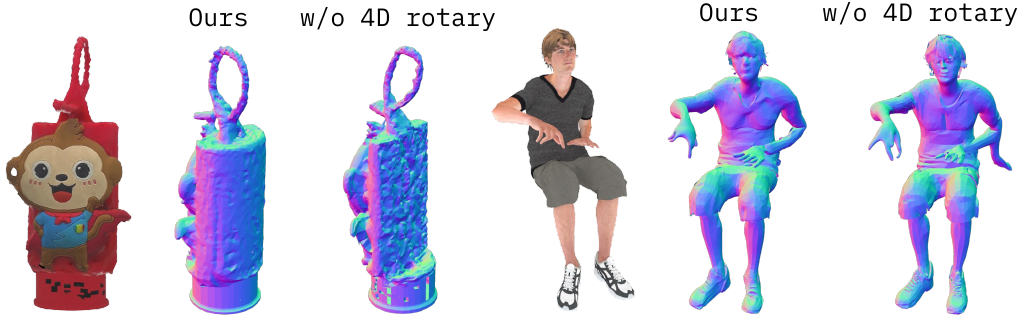}
    \captionof{figure}{\textbf{Qualitative comparisons of the 4D rotary embedding.}
    Compared with our full model, removing 4D rotary leads to degraded geometry, including rough back surfaces and extra arms.}
    \label{fig:results_abla}
  \end{minipage}
\vspace{-0.5cm}
\end{figure}
We analyze each design choice by removing one component at a time from the second stage (mesh) model. Results on our held-out TexVerse-1k benchmark are reported in~\tabref{tab:ablation}. 
%
We observe that all three components contribute to the generation quality. The drop is largest when the first-frame conditioning is removed, consistent with our motivation that the first-frame anchor makes the 4D generation task easier. Removing 4D rotary embedding by replacing it with 1D temporal RoPE applied on top of the 3D spatial RoPE causes the model to confuse tokens at different time steps, and this hurts both geometric and semantic accuracy. 

The numerical gap on this row is modest because we believe the metrics saturate, but visual inspection in \figref{fig:results_abla} shows the cost clearly: without the 4D rotary embedding, reconstructions exhibit an extra arm and a rough back surface, whereas our model produces clean geometry and correct articulation. The \emph{w/o anchor attention} row replaces our sliding-window-plus-anchor pattern with full self-attention. Full attention has too much unnecessary context, while restricting attention to a local window plus a global first-frame anchor matches the structure of the task: short-range motion is captured locally, and the anchor maintains a consistent global identity. 

Next, we compare five attention patterns: \emph{full} attention across all $F$ frames, \emph{causal} attention, where frame $f$ attends only to frame $1, \dots, f$, \emph{sliding} window, where frame $f$ attends to $[f - w/2, f + w/2]$,  
\vspace{-0.58cm}

\begin{wraptable}[10]{r}{0.6\textwidth}
  \centering
  \vspace{-0.35cm}
  \caption{\textbf{Cross-frame attention pattern.} Time is wall-clock normalized to ours. Sliding window + anchor (ours) is best on every metric while running $2.3\times$ faster than full attention.}
  \label{tab:ablation_attention}
  \vspace{-0.1cm}
  \small
      \resizebox{\linewidth}{!}{
  \begin{tabular}{lcccccc}
    \toprule
     & \textbf{CD-3D} $\downarrow$ & \textbf{CD-4D} $\downarrow$ & \textbf{ULIP-2} $\uparrow$ & \textbf{Uni3D} $\uparrow$ & \textbf{Time} $\downarrow$ \\
    \midrule
    Full attention                      & 0.0470 & 0.1009 & 0.2835 & 0.2447 & $2.3\times$ \\
    Causal  attention                   & 0.0491 & 0.1038 &0.2838 & 0.2564 &  $1.5\times$ \\
    Sliding window             & 0.0468 & 0.1085 & 0.2870 & 0.2538 &  $0.9\times$ \\
    Spatial   attention                 &  0.0471 & 0.1011& 0.2791 & 0.2548 &  $\textbf{0.4}\times$ \\
    \midrule
    Ours                       & \textbf{0.0454} & \textbf{0.0970} & \textbf{0.2879} & \textbf{0.2576} &  $1.0\times$ \\
    \bottomrule
  \end{tabular}
  }
\end{wraptable}
spatial,
where $N^3$ voxel grid is partitioned into $8 \times 8 \times 8$ blocks 
and tokens attend across time only within their block, 
and our is \emph{sliding window + anchor}, which augments the local window with all tokens from the first frame.
\tabref{tab:ablation_attention} shows that our pattern achieves the best
score on all four quality metrics while remaining competitive in cost. Full
attention is the most expensive yet underperforms ours,
indicating that full attention injects noise rather than
than a useful signal. A plain sliding window gives the worst CD-4D, since each frame can drift away from the initial geometry without a stable global reference. Spatial attention is the cheapest overall but caps quality because tokens cannot exchange information across blocks. Our sliding window\,+\,anchor combines the
locality of sliding window with a single global anchor, recovering temporal
consistency.

\section{Conclusion}
We presented \methodname{}, an approach that adapts a pretrained image-to-3D generator into a video-to-4D mesh generator. We introduce a lightweight conversion of Trellis2 to dynamic generation through temporal ReRoPE, sliding-window cross-frame attention with an anchor frame, and first-frame conditioning. This design preserves the strong geometry and material prior of O-Voxels while adding temporal consistency, allowing the model to handle topology changes, transparent objects, thin structures, and inner surfaces. Experiments show strong geometric accuracy on ActionBench and consistent gains over recent video-to-4D baselines. This is important for the community because it shows that strong static 3D foundation models can be lifted to 4D without training a new model from scratch. Future work aims to extend \methodname{} to longer videos, scene-level interactions, stronger camera motion, and more accurate physical dynamics.

 \section{Acknowledgment}
We thank Ashkan Mirzaei for valuable discussions, and Vladislav Shakhray and Gleb Dmukhin for their help with Blender.

{
\small
\bibliographystyle{ieee_fullname}
\bibliography{ref}

@String(CVPR= {IEEE Conf. Comput. Vis. Pattern Recog.})

@String(ICCV= {Int. Conf. Comput. Vis.})

@String(ECCV= {Eur. Conf. Comput. Vis.})

@String(TOG= {ACM Trans. Graph.})

@String(ICLR = {Int. Conf. Learn. Represent.})

@String(CVPR  = {CVPR})

@String(ICCV  = {ICCV})

@String(ECCV  = {ECCV})

@String(TOG   = {ACM TOG})

@String(ICLR  = {ICLR})

@article{trellis2,
  title={Native and Compact Structured Latents for {{3D}} Generation},
  author={Xiang, Jianfeng and Chen, Xiaoxue and Xu, Sicheng and Wang, Ruicheng and Lv, Zelong and Deng, Yu and Zhu, Hongyuan and Dong, Yue and Zhao, Hao and Yuan, Nicholas Jing and others},
  journal={arXiv preprint arXiv:2512.14692},
  year={2025}
}

@article{ss4d,
  title={{{SS4D}}: Native {{4D}} Generative Model via Structured Spacetime Latents},
  author={Li, Zhibing and Zhang, Mengchen and Wu, Tong and Tan, Jing and Wang, Jiaqi and Lin, Dahua},
  journal={TOG},
  year={2025}
}

@inproceedings{shapegen4d,
  title={{{ShapeGen4D}}: Towards High Quality {{4D}} Shape Generation from Videos},
  author={Yenphraphai, Jiraphon and Mirzaei, Ashkan and Chen, Jianqi and Zou, Jiaxu and Tulyakov, Sergey and Yeh, Raymond A and Wonka, Peter and Wang, Chaoyang},
  booktitle={Proc. ICLR},
  year={2026}
}

@inproceedings{actionmesh,
  title={{{ActionMesh}}: Animated {{3D}} Mesh Generation with Temporal {{3D}} Diffusion},
  author={Sabathier, Remy and Novotny, David and Mitra, Niloy J and Monnier, Tom},
  booktitle={Proc. CVPR},
  year={2026}
}

@article{mesh4d,
  title={{{Mesh4D}}: {{4D}} Mesh Reconstruction and Tracking from Monocular Video},
  author={Jiang, Zeren and Zheng, Chuanxia and Laina, Iro and Larlus, Diane and Vedaldi, Andrea},
  journal={arXiv preprint arXiv:2601.05251},
  year={2026}
}

@article{motion324,
  title={Motion 3-to-4: {{3D}} Motion Reconstruction for {{4D}} Synthesis},
  author={Chen, Hongyuan and Chen, Xingyu and Zhang, Youjia and Xu, Zexiang and Chen, Anpei},
  journal={arXiv preprint arXiv:2601.14253},
  year={2026}
}

@article{zhao2025hunyuan3d,
  title={{{Hunyuan3D}} 2.0: Scaling Diffusion Models for High Resolution Textured {{3D}} Assets Generation},
  author={Zhao, Zibo and Lai, Zeqiang and Lin, Qingxiang and Zhao, Yunfei and Liu, Haolin and Yang, Shuhui and Feng, Yifei and Yang, Mingxin and Zhang, Sheng and Yang, Xianghui and others},
  journal={arXiv preprint arXiv:2501.12202},
  year={2025}
}

@article{li2025step1x,
  title={{{Step1X-3D}}: Towards High-Fidelity and Controllable Generation of Textured {{3D}} Assets},
  author={Li, Weiyu and Zhang, Xuanyang and Sun, Zheng and Qi, Di and Li, Hao and Cheng, Wei and Cai, Weiwei and Wu, Shihao and Liu, Jiarui and Wang, Zihao and others},
  journal={arXiv preprint arXiv:2505.07747},
  year={2025}
}

@inproceedings{xiang2024structured,
  title={Structured {{3D}} Latents for Scalable and Versatile {{3D}} Generation},
  author={Xiang, Jianfeng and Lv, Zelong and Xu, Sicheng and Deng, Yu and Wang, Ruicheng and Zhang, Bowen and Chen, Dong and Tong, Xin and Yang, Jiaolong},
  booktitle={Proc. CVPR},
  year={2025}
}

@article{rope,
  title={{{RoFormer}}: Enhanced Transformer with Rotary Position Embedding},
  author={Su, Jianlin and Ahmed, Murtadha and Lu, Yu and Pan, Shengfeng and Bo, Wen and Liu, Yunfeng},
  journal={Neurocomputing},
  year={2024}
}

@article{rerope,
  title={{{ReRoPE}}: Repurposing {{RoPE}} for Relative Camera Control},
  author={Li, Chunyang and Yang, Yuanbo and Shao, Jiahao and Zhou, Hongyu and Schwarz, Katja and Liao, Yiyi},
  journal={arXiv preprint arXiv:2602.08068},
  year={2026}
}

@article{texverse,
  title={{{Texverse}}: A universe of {{3D}} objects with high-resolution textures},
  author={Zhang, Yibo and Zhang, Li and Ma, Rui and Cao, Nan},
  journal={arXiv preprint arXiv:2508.10868},
  year={2025}
}

@inproceedings{radfordclip,
  title={Learning transferable visual models from natural language supervision},
  author={Radford, Alec and Kim, Jong Wook and Hallacy, Chris and Ramesh, Aditya and Goh, Gabriel and Agarwal, Sandhini and Sastry, Girish and Askell, Amanda and Mishkin, Pamela and Clark, Jack and others},
  booktitle={Proc. ICML},
  year={2021}
}

@inproceedings{xue2023ulip,
  title={{{ULIP}}: Learning a unified representation of language, images, and point clouds for {{3D}} understanding},
  author={Xue, Le and Gao, Mingfei and Xing, Chen and Mart{\'\i}n-Mart{\'\i}n, Roberto and Wu, Jiajun and Xiong, Caiming and Xu, Ran and Niebles, Juan Carlos and Savarese, Silvio},
  booktitle={Proc. CVPR},
  year={2023}
}

@article{zhou2023uni3d,
  title={{{Uni3D}}: Exploring unified {{3D}} representation at scale},
  author={Zhou, Junsheng and Wang, Jinsheng and Ma, Baorui and Liu, Yu-Shen and Huang, Tiejun and Wang, Xinlong},
  journal={arXiv preprint arXiv:2310.06773},
  year={2023}
}

@article{unterthiner2019fvd,
  title={{{FVD}}: A new metric for video generation},
  author={Unterthiner, Thomas and Van Steenkiste, Sjoerd and Kurach, Karol and Marinier, Rapha{\"e}l and Michalski, Marcin and Gelly, Sylvain},
  year={2019}
}

@article{fu2023dreamsim,
  title={{{DreamSim}}: Learning new dimensions of human visual similarity using synthetic data},
  author={Fu, Stephanie and Tamir, Netanel and Sundaram, Shobhita and Chai, Lucy and Zhang, Richard and Dekel, Tali and Isola, Phillip},
  journal={arXiv preprint arXiv:2306.09344},
  year={2023}
}

@article{loshchilov2017adamw,
  title={Decoupled weight decay regularization},
  author={Loshchilov, Ilya and Hutter, Frank},
  journal={arXiv preprint arXiv:1711.05101},
  year={2017}
}

@inproceedings{zheng2024unified,
  title={A Unified Approach for Text- and Image-Guided {{4D}} Scene Generation},
  author={Zheng, Yufeng and Li, Xueting and Nagano, Koki and Liu, Sifei and Hilliges, Otmar and De Mello, Shalini},
  booktitle={Proc. CVPR},
  year={2024}
}

@inproceedings{ling2024,
  title={Align Your {{Gaussians}}: Text-to-{{4D}} with Dynamic {{3D}} {{Gaussians}} and Composed Diffusion Models},
  author={Ling, Huan and Kim, Seung Wook and Torralba, Antonio and Fidler, Sanja and Kreis, Karsten},
  booktitle={Proc. CVPR},
  year={2024}
}

@inproceedings{Yuan_2024_CVPR,
  title={{{GAvatar}}: Animatable {{3D}} {{Gaussian}} Avatars with Implicit Mesh Learning},
  author={Yuan, Ye and Li, Xueting and Huang, Yangyi and De Mello, Shalini and Nagano, Koki and Kautz, Jan and Iqbal, Umar},
  booktitle={Proc. CVPR},
  year={2024}
}

@inproceedings{bahmani2024,
  title={{{4D-fy}}: Text-to-{{4D}} Generation Using Hybrid Score Distillation Sampling},
  author={Bahmani, Sherwin and Skorokhodov, Ivan and Rong, Victor and Wetzstein, Gordon and Guibas, Leonidas and Wonka, Peter and Tulyakov, Sergey and Park, Jeong Joon and Tagliasacchi, Andrea and Lindell, David B.},
  booktitle={Proc. CVPR},
  year={2024}
}

@inproceedings{wu2024cat4d,
  title={{{CAT4D}}: Create Anything in {{4D}} with Multi-View Video Diffusion Models},
  author={Wu, Rundi and Gao, Ruiqi and Poole, Ben and Trevithick, Alex and Zheng, Changxi and Barron, Jonathan T. and Holynski, Aleksander},
  booktitle={Proc. CVPR},
  year={2025}
}

@article{gvfd,
  title={{{Gaussian}} Variation Field Diffusion for High-Fidelity Video-to-{{4D}} Synthesis},
  author={Zhang, Bowen and Xu, Sicheng and Wang, Chuxin and Yang, Jiaolong and Zhao, Feng and Chen, Dong and Guo, Baining},
  journal={arXiv preprint arXiv:2507.23785},
  year={2025}
}

@inproceedings{sun2024dimensionx,
  title={{{DimensionX}}: Create Any {{3D}} and {{4D}} Scenes from a Single Image with Decoupled Video Diffusion},
  author={Sun, Wenqiang and Chen, Shuo and Liu, Fangfu and Chen, Zilong and Duan, Yueqi and Zhu, Jun and Zhang, Jun and Wang, Yikai},
  booktitle={Proc. ICCV},
  year={2025}
}

@inproceedings{yao2025sv4d,
  title={{{SV4D}} 2.0: Enhancing Spatio-Temporal Consistency in Multi-View Video Diffusion for High-Quality {{4D}} Generation},
  author={Yao, Chun-Han and Xie, Yiming and Voleti, Vikram and Jiang, Huaizu and Jampani, Varun},
  booktitle={Proc. ICCV},
  year={2025}
}

@inproceedings{liu2025free4d,
  title={{{Free4D}}: Tuning-Free {{4D}} Scene Generation with Spatial-Temporal Consistency},
  author={Liu, Tianqi and Huang, Zihao and Chen, Zhaoxi and Wang, Guangcong and Hu, Shoukang and Shen, Liao and Sun, Huiqiang and Cao, Zhiguo and Li, Wei and Liu, Ziwei},
  booktitle={Proc. ICCV},
  year={2025}
}

@inproceedings{jin2025diffuman4d,
  title={{{Diffuman4D}}: {{4D}} Consistent Human View Synthesis from Sparse-View Videos with Spatio-Temporal Diffusion Models},
  author={Jin, Yudong and Peng, Sida and Wang, Xuan and Xie, Tao and Xu, Zhen and Yang, Yifan and Shen, Yujun and Bao, Hujun and Zhou, Xiaowei},
  booktitle={Proc. ICCV},
  year={2025}
}

@article{wan2025,
  title={Wan: Open and Advanced Large-Scale Video Generative Models}, 
  author={Team Wan and Ang Wang and Baole Ai and Bin Wen and Chaojie Mao and Chen-Wei Xie and Di Chen and Feiwu Yu and Haiming Zhao and Jianxiao Yang and Jianyuan Zeng and Jiayu Wang and Jingfeng Zhang and Jingren Zhou and Jinkai Wang and Jixuan Chen and Kai Zhu and Kang Zhao and Keyu Yan and Lianghua Huang and Mengyang Feng and Ningyi Zhang and Pandeng Li and Pingyu Wu and Ruihang Chu and Ruili Feng and Shiwei Zhang and Siyang Sun and Tao Fang and Tianxing Wang and Tianyi Gui and Tingyu Weng and Tong Shen and Wei Lin and Wei Wang and Wei Wang and Wenmeng Zhou and Wente Wang and Wenting Shen and Wenyuan Yu and Xianzhong Shi and Xiaoming Huang and Xin Xu and Yan Kou and Yangyu Lv and Yifei Li and Yijing Liu and Yiming Wang and Yingya Zhang and Yitong Huang and Yong Li and You Wu and Yu Liu and Yulin Pan and Yun Zheng and Yuntao Hong and Yupeng Shi and Yutong Feng and Zeyinzi Jiang and Zhen Han and Zhi-Fan Wu and Ziyu Liu},
  journal={arXiv preprint arXiv:2503.20314},
  year={2025}
}

@inproceedings{ren2024l4gm,
  title={{{L4GM}}: Large {{4D}} {{Gaussian}} reconstruction model},
  author={Ren, Jiawei and Xie, Kevin and Mirzaei, Ashkan and Liang, Hanxue and Zeng, Xiaohui and Kreis, Karsten and Liu, Ziwei and Torralba, Antonio and Fidler, Sanja and Kim, Seung W and others},
  booktitle={Proc. NeurIPS},
  year={2024}
}

@article{yin2026sculpt4d,
  title={{{Sculpt4D}}: Generating {{4D}} Shapes via Sparse-Attention Diffusion Transformers},
  author={Yin, Minghao and Hu, Wenbo and Xu, Jiale and Shan, Ying and Han, Kai},
  journal={arXiv preprint arXiv:2604.21592},
  year={2026}
}

@article{poole2022dreamfusion,
  title={{{DreamFusion}}: Text-to-{{3D}} using {{2D}} diffusion},
  author={Poole, Ben and Jain, Ajay and Barron, Jonathan T and Mildenhall, Ben},
  journal={arXiv preprint arXiv:2209.14988},
  year={2022}
}

@article{singer2023text,
  title={Text-to-{{4D}} dynamic scene generation},
  author={Singer, Uriel and Sheynin, Shelly and Polyak, Adam and Ashual, Oron and Makarov, Iurii and Kokkinos, Filippos and Goyal, Naman and Vedaldi, Andrea and Parikh, Devi and Johnson, Justin and others},
  journal={arXiv preprint arXiv:2301.11280},
  year={2023}
}

@article{jiang2023consistent4d,
  title={{{Consistent4D}}: Consistent 360° dynamic object generation from monocular video},
  author={Jiang, Yanqin and Zhang, Li and Gao, Jin and Hu, Weimin and Yao, Yao},
  journal={arXiv preprint arXiv:2311.02848},
  year={2023}
}

@article{ren2023dreamgaussian4d,
  title={{{DreamGaussian4D}}: Generative {{4D}} gaussian splatting},
  author={Ren, Jiawei and Pan, Liang and Tang, Jiaxiang and Zhang, Chi and Cao, Ang and Zeng, Gang and Liu, Ziwei},
  journal={arXiv preprint arXiv:2312.17142},
  year={2023}
}

@article{yin20234dgen,
  title={{{4DGen}}: Grounded {{4D}} content generation with spatial-temporal consistency},
  author={Yin, Yuyang and Xu, Dejia and Wang, Zhangyang and Zhao, Yao and Wei, Yunchao},
  journal={arXiv preprint arXiv:2312.17225},
  year={2023}
}

@inproceedings{zeng2024stag4d,
  title={{{STAG4D}}: Spatial-temporal anchored generative {{4D}} gaussians},
  author={Zeng, Yifei and Jiang, Yanqin and Zhu, Siyu and Lu, Yuanxun and Lin, Youtian and Zhu, Hao and Hu, Weiming and Cao, Xun and Yao, Yao},
  booktitle={Proc. ECCV},
  year={2024}
}

@inproceedings{wang2024vidu4d,
  title={{{Vidu4D}}: Single generated video to high-fidelity {{4D}} reconstruction with dynamic gaussian surfels},
  author={Wang, Yikai and Wang, Xinzhou and Chen, Zilong and Wang, Zhengyi and Sun, Fuchun and Zhu, Jun},
  booktitle={Proc. NeurIPS},
  year={2024}
}

@inproceedings{zhang20244diffusion,
  title={{{4Diffusion}}: Multi-view video diffusion model for {{4D}} generation},
  author={Zhang, Haiyu and Chen, Xinyuan and Wang, Yaohui and Liu, Xihui and Wang, Yunhong and Qiao, Yu},
  booktitle={Proc. NeurIPS},
  year={2024}
}

@inproceedings{chen2025v2m4,
  title={{{V2M4}}: {{4D}} mesh animation reconstruction from a single monocular video},
  author={Chen, Jianqi and Zhang, Biao and Tang, Xiangjun and Wonka, Peter},
  booktitle={Proc. ICCV},
  year={2025}
}

@article{zhang20233dshape2vecset,
  title={{{3DShape2VecSet}}: A {{3D}} shape representation for neural fields and generative diffusion models},
  author={Zhang, Biao and Tang, Jiapeng and Niessner, Matthias and Wonka, Peter},
  journal={TOG},
  year={2023}
}

@article{li2025triposg,
  title={{{TripoSG}}: High-fidelity {{3D}} shape synthesis using large-scale rectified flow models},
  author={Li, Yangguang and Zou, Zi-Xin and Liu, Zexiang and Wang, Dehu and Liang, Yuan and Yu, Zhipeng and Liu, Xingchao and Guo, Yuan-Chen and Liang, Ding and Ouyang, Wanli and others},
  journal={TPAMI},
  year={2025}
}

@article{hunyuan3d2025hunyuan3d,
  title={{{Hunyuan3D}} 2.1: From images to high-fidelity {{3D}} assets with production-ready {{PBR}} material},
  author={Hunyuan3D, Team and Yang, Shuhui and Yang, Mingxin and Feng, Yifei and Huang, Xin and Zhang, Sheng and He, Zebin and Luo, Di and Liu, Haolin and Zhao, Yunfei and others},
  journal={arXiv preprint arXiv:2506.15442},
  year={2025}
}

@article{lai2025hunyuan3d,
  title={{{Hunyuan3D}} 2.5: Towards high-fidelity {{3D}} assets generation with ultimate details},
  author={Lai, Zeqiang and Zhao, Yunfei and Liu, Haolin and Zhao, Zibo and Lin, Qingxiang and Shi, Huiwen and Yang, Xianghui and Yang, Mingxin and Yang, Shuhui and Feng, Yifei and others},
  journal={arXiv preprint arXiv:2506.16504},
  year={2025}
}

@article{wu2025direct3d,
  title={{{Direct3D-S2}}: Gigascale {{3D}} generation made easy with spatial sparse attention},
  author={Wu, Shuang and Lin, Youtian and Zhang, Feihu and Zeng, Yifei and Yang, Yikang and Bao, Yajie and Qian, Jiachen and Zhu, Siyu and Cao, Xun and Torr, Philip and others},
  journal={arXiv preprint arXiv:2505.17412},
  year={2025}
}

@inproceedings{trellis,
  title={Structured {{3D}} latents for scalable and versatile {{3D}} generation},
  author={Xiang, Jianfeng and Lv, Zelong and Xu, Sicheng and Deng, Yu and Wang, Ruicheng and Zhang, Bowen and Chen, Dong and Tong, Xin and Yang, Jiaolong},
  booktitle={Proc. CVPR},
  year={2025}
}
}

\clearpage
\appendix

\setcounter{section}{0}
\renewcommand{\theHsection}{A\arabic{section}}
\renewcommand{\thesection}{A\arabic{section}}
\renewcommand{\thetable}{A\arabic{table}}
\setcounter{table}{0}
\setcounter{figure}{0}
\renewcommand{\thetable}{A\arabic{table}}
\renewcommand\thefigure{A\arabic{figure}}
\renewcommand{\theHtable}{A.Tab.\arabic{table}}
\renewcommand{\theHfigure}{A.Abb.\arabic{figure}}
\renewcommand\theequation{A\arabic{equation}}
\renewcommand{\theHequation}{A.Abb.\arabic{equation}}

{\centering \Large \textbf{Appendix}}

\begin{table}[H]
  \centering
  \caption{\textbf{Ablation on the RoPE ratio.} 
We evaluate on 32 held-out objects from TexVerse~\cite{texverse}. We use the full spatial RoPE setting (ratio $=1.0$) as the reference and report the CD-3D between the mesh generated by each ratio and this reference.}
  \label{tab:rerope_ratio}
  \small
  \setlength{\tabcolsep}{8pt}
  \renewcommand{\arraystretch}{1.2}
  \begin{tabular}{lcccccc}
    \toprule
    & \textbf{0.0} & \textbf{0.2} & \textbf{0.4} & \textbf{0.6} & \textbf{0.8} & \textbf{1.0} \\
    \midrule
    CD-3D $\downarrow$ & 0.0181 & 0.0098 & 0.0079 & 0.0079 & 0.0079 & 0 \\
    \bottomrule
  \end{tabular}
\end{table}
\section{Limitations or failure cases}
\label{sec:limitation}
Our method inherits two limitations from the Trellis2 backbone we built upon. First, the generated meshes occasionally contain holes, since Trellis2 does not enforce a watertight-mesh assumption and provides no guarantee that the output surface will be closed. Second, the textures produced by Trellis2 exhibit a color-shifting artifact, which could result in the generation of metallic surfaces for transparent objects.

Our own design introduces another limitation: because we output a mesh sequence rather than a single static mesh, the results sometimes lack temporal consistency, particularly in regions with high-frequency geometric details.

\section{User Study Details}
\label{sec:user_study_details}

We conduct a user study to evaluate the perceived quality of our generated dynamic 3D reconstructions against prior methods. Participants compare pairwise rendered videos and select which result looks better overall, considering visual quality, temporal consistency, and preservation of details.

\subsection{Instructions Shown to Participants}

Participants were shown the following task description:

\begin{quote}
\small
\textbf{4D Reconstruction.}
In this study, you will compare pairs of rendered 4D reconstruction results.
Each question shows a single video containing: an input video, Result A, and Result B.
Your task is to decide which generated result, A or B, looks better overall.
\end{quote}

Participants were instructed to consider the following criteria:

\begin{enumerate}
    \item \textbf{Overall visual quality:} which result looks more realistic and visually appealing.
    \item \textbf{Temporal consistency:} which result has smoother and more coherent motion over time.
    \item \textbf{Preservation of details:} which result better preserves fine details.
\end{enumerate}

Each question used the following prompt:

\begin{quote}
\small
Watch the input video and the two generated results, A and B. Which result looks better overall?
\end{quote}

The available answer choices were:
\begin{itemize}
    \item Result A is better.
    \item Result B is better.
    \item Tie / no clear difference.
\end{itemize}

Participants were told that there were 25 questions in total, that they could go back to change previous answers, and that the study would take approximately 10 minutes.

\subsection{Participants and Comparisons}

\begin{table}[h]
\centering
\small
\caption{\textbf{User study summary.}}
\label{tab:user_study_summary}
\begin{tabular}{lc}
\toprule
\textbf{Statistic} & \textbf{Value} \\
\midrule
Total unique participant IDs & 21 \\
Participants completing all 25 questions & 18 \\
Partial participants & 3 \\
Total response rows collected & 458 \\
Questions per participant & 25 \\
RGB comparisons & 15 \\
Normal-map comparisons & 10 \\
\bottomrule
\end{tabular}
\end{table}

The study contains 25 pairwise comparison questions per participant. Among these, 15 questions evaluate RGB rendered appearance, while 10 questions evaluate normal-map renderings that emphasize surface geometry. In the RGB category, we compare against Motion 3-to-4~\cite{motion324}, Mesh4D~\cite{mesh4d}, and SS4D~\cite{ss4d}. In the normal-map category, we compare against ActionMesh~\cite{actionmesh} and ShapeGen4D~\cite{shapegen4d}.
The question order was randomized independently for each participant. Responses were collected anonymously.

We compute the win rate as follows:
\begin{equation}
    \mathrm{WinRate} =
    \frac{W + 0.5T}{N}
\end{equation}
where $W, L, N$ are the number of wins, losses, and ties for our method. 
\begin{figure}[t]
  \centering
  \includegraphics[width=0.8\linewidth]{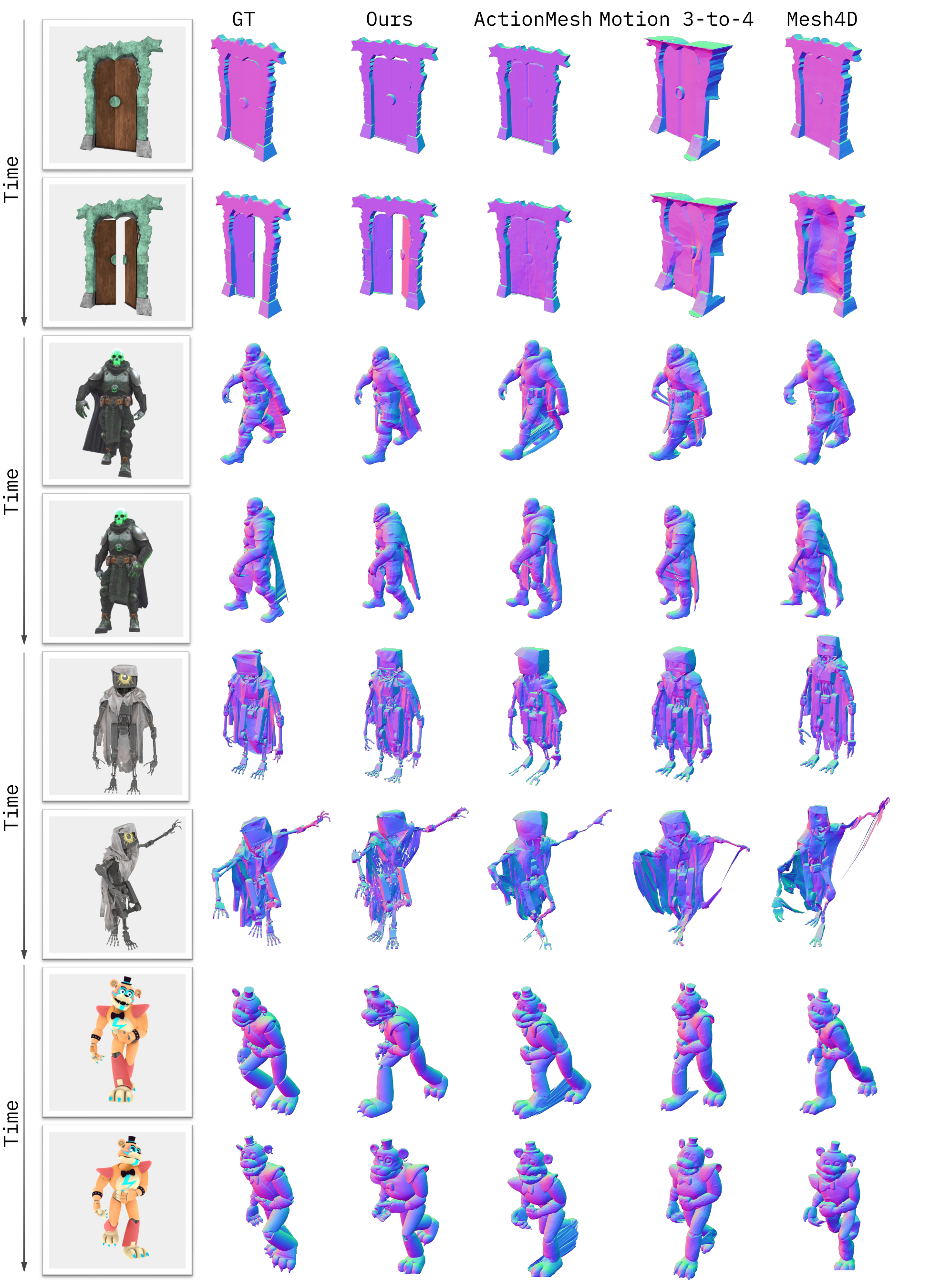}
  \caption{\textbf{Qualitative comparison on TexVerse test set.} The left column shows input video frames. Each pair of rows shows two frames from the generated sequence, with normal renderings from ground truth and each method. Baselines (ActionMesh~\cite{actionmesh}, Motion 3-to-4~\cite{motion324}, Mesh4D~\cite{mesh4d}) struggle with topology changes: the opening door fuses shut, character limbs and capes stick to the body, and motion fails to propagate to the second frame. Our method follows the ground-truth motion and preserves fine geometry across frames.}
  \label{fig:results_complex}
\end{figure}


\section{Helix4DBench Construction}
\label{sec:benchmark_construction}

Our test set is constructed from the example images released in the official Trellis2 repository. We use each image as a static object input and generate a video using Wan2.2~\cite{wan2025} with the corresponding text prompt in \tabref{tab:benchmark_prompts}. Each prompt shares the same prefix:
\begin{quote}
\small
\emph{Fixed camera, pure white background, no camera movement or rotation.}
\end{quote}
We therefore omit this shared prefix from the table for compactness and report only the motion-specific prompt suffix. For each image filename, we report only its first eight characters, i.e., \texttt{filename[:8]}; for the special filename \texttt{1.png}, we report the prefix as \texttt{1}.

\begin{longtable}{p{0.14\linewidth}p{0.78\linewidth}}
\caption{\textbf{Image IDs and their prompts.} The image ID column shows the filename prefix which is the first eight characters of the original image filename.}
\label{tab:benchmark_prompts}\\
\toprule
\textbf{Image ID} & \textbf{Prompt suffix} \\
\midrule
\endfirsthead
\toprule
\textbf{Image ID} & \textbf{Prompt suffix} \\
\midrule
\endhead

\texttt{2bb09323} & A steampunk mechanical pig statue where the rock base stays fixed while the pig springs to life, galloping vigorously in place with large exaggerated leg strides, pistons pumping hard, steam bursting from joints, and glowing parts pulsing brightly. \\

\texttt{4dae7ef0} & A medieval wooden catapult where the base stays grounded while the throwing arm swings forcefully upward in a full arc, launching a projectile, with chains whipping and ropes snapping taut with dramatic motion. \\

\texttt{5c80e5e0} & A fierce bear rearing up on its hind legs, swiping both front claws aggressively through the air, roaring with its jaw wide open, and twisting its torso side to side with powerful exaggerated motion. \\

\texttt{5a6c81d3} & A detailed hiking backpack where zippers rapidly unzip, multiple compartments swing open wide, straps flap loosely, and the whole bag bounces as if being shaken out and unpacked with energetic motion. \\

\texttt{7b540da3} & A robed woman turning her full torso dramatically from side to side, arms sweeping outward, robes billowing and flowing with wide expressive gestures and dynamic cloth movement. \\

\texttt{8ce83f6a} & A stylized mermaid swaying her torso and arms expressively while her long hair flows and whips around, as glowing jellyfish rise and drift around her with large undulating tentacle motion. \\

\texttt{7d7659d5} & A mechanical suit of armor raising both arms high, clenching and unclenching its fists, twisting its torso left and right, with every joint visibly bending and articulating with bold mechanical motion. \\

\texttt{8e12cf09} & A retro mechanical robot waving both arms up and down, bending at the waist, turning its head left and right, with gears spinning and joints clicking in large expressive movements. The viewpoint does not change. \\

\texttt{7bd0521d} & A grotesque goblin bust where the face contorts dramatically, jaw dropping wide open, tongue thrashing out, eyes bulging, and the head tilting and shaking with exaggerated snarling facial motion. \\

\texttt{9c306c7b} & A stylized traveler marching energetically in place with high knee lifts, arms swinging wide, backpack bouncing, and gear swaying with lively exaggerated walking motion. \\

\texttt{50b70c5f} & A Viking warrior swinging his axe in a wide arc overhead, the lantern swaying broadly on its chain, his torso twisting with the swing, cape and armor shifting with strong dynamic motion. \\

\texttt{61fea9d0} & A stylized archer pulling the bowstring all the way back with full arm extension, torso twisting into the draw, then releasing the arrow with a dramatic snap and follow-through motion. \\

\texttt{80ad7988} & A stylized armored figure raising the golden chalice high overhead with a full arm lift, then bringing it down and tilting it forward as if pouring, with the other arm gesturing outward expressively. \\

\texttt{95db3c13} & A quadruped battle robot shifting its weight between legs, turret swiveling rapidly back and forth, legs stomping and repositioning with heavy mechanical steps and visible hydraulic motion. \\

\texttt{154c8867} & A stylized potted plant where stems shoot upward rapidly, leaves unfurl and spread wide, colorful flowers burst into full bloom, and vines curl and extend outward with fast energetic growing motion. \\

\texttt{454e7d8a} & A stylized human lifting and turning the small robot in his hands, the robot flailing its arms and kicking its legs, while the human leans and shifts his body to keep hold of it with lively playful motion. \\

\texttt{901d8de4} & A glass bottle aquarium where fish dart and chase each other in fast circles, bubbles stream upward rapidly, and aquatic plants sway dramatically with strong swirling water motion inside the bottle. \\

\texttt{3903b879} & A humanoid robot marching in place with high exaggerated steps, swinging its arms wide, turning its torso and head dramatically from side to side with bold mechanical motion. \\

\texttt{26717a7d} & A large tropical tree where branches burst outward rapidly, leaves unfurl and flutter vigorously in the wind, vines whip and spiral dynamically, and the entire canopy sways with strong dramatic motion. \\

\texttt{39488b45} & A marble bust where the face dramatically comes to life, eyes widening, mouth opening and closing as if speaking passionately, head turning left and right with bold expressive facial motion. \\

\texttt{52284bf4} & A marble angel statue where the wings spread wide open in a dramatic full extension, arms lifting outward, and the entire upper body shifting into a new majestic pose with sweeping lifelike motion. \\

\texttt{65433d02} & A vintage typewriter where keys hammer down rapidly in quick succession, the carriage slides across with a snap, the return lever slams back, and paper feeds upward fast with energetic clacking motion. \\

\texttt{3723615e} & A stylized witch leaping high into the air, arms spread wide, hat flying off her head, hair whipping upward, hanging bottles and charms swinging wildly with exaggerated bouncy motion. \\

\texttt{a13d176c} & An ornate steampunk pistol where gears spin rapidly, the hammer cocks back and snaps forward, the slide racks with a sharp motion, and small valves and pistons pump with fast visible mechanical action. \\

\texttt{1} & A steampunk mechanical device where gears spin rapidly at different speeds, pistons pump hard with visible force, glowing fluid rushes through transparent tubes, and steam bursts from valves with energetic continuous motion. \\

\texttt{f94e2b76} & A stylized turkey character with a jetpack performing a playful leap upward, legs tucking in, wings and feathers bouncing wildly, and the jetpack flaring with bright exhaust bursts and lively energetic motion. \\

\texttt{f5332118} & A stone king statue rising forcefully from a seated throne to a full standing position, gripping his sword tightly, with heavy weighty motion, stone cracking and shifting, and dust particles falling off. \\

\texttt{e4d6b2f3} & A stylized rider on a stationary motorcycle turning his torso toward the viewer and raising a hand high to wave enthusiastically with bold upper-body motion and natural arm swing. \\

\texttt{e1046572} & A stylized farmer dropping a basket from his hands, the basket tumbling downward, while his facial expression shifts dramatically from calm to wide-eyed surprise with exaggerated body recoil. \\

\texttt{cdf996a6} & A fierce goblin warrior swinging its weapon in wide aggressive arcs, torso twisting with each swing, feet shifting for balance, with powerful fluid slashing motion and visible follow-through. \\

\texttt{c3d714bc} & A monstrous treasure chest where the base stays still while the massive mouth slowly closes with sharp teeth interlocking, the tongue retracting inward, and drool dripping with smooth organic motion. \\

\texttt{c2125d08} & A wooden windmill where the structure remains still while the turbine blades spin smoothly and continuously with steady natural motion and subtle creaking sway. \\

\texttt{6b6d89d4} & A pair of rugged leather boots performing a rhythmic tango-style tap dance in place with precise energetic footwork, heels clicking, toes tapping, and bouncing with lively motion. \\

\texttt{0e4984a9} & A brick wall where leaves and tendrils rapidly grow and spread outward, curling and expanding to fully cover the surface with fast organic creeping motion and unfurling vines. \\

\texttt{290af2dd} & A golden armored helmet where the metal slowly melts and deforms downward with soft fluid motion, edges dripping and pooling naturally, gold surface warping and sagging dramatically. \\

\texttt{ab3bb3e1} & An hourglass where sand pours smoothly and continuously from the top chamber to the bottom, grains cascading and piling up with visible flowing motion. \\

\texttt{c9340e74} & A bat perched on a tomb lifting off and flapping its wings vigorously to fly upward with strong wing beats, body rising and legs releasing their grip with dramatic takeoff motion. \\

\texttt{0f168a4b} & A mounted turret firing rapid glowing energy shots forward, the upper gun assembly swiveling and tracking targets, muzzle flashing brightly, with the base remaining completely static. \\

\texttt{25d412fe} & Potion bottles shattering and exploding outward with colorful liquid splashing in all directions, glass shards flying, and vibrant fluid arcing through the air with dramatic explosive motion. \\

\texttt{7baa867b} & An ornate chair tipping over sideways and crashing to the ground with a heavy fall, legs bouncing on impact, and the whole frame rocking with dramatic toppling motion. \\

\texttt{bb319089} & A truck house where doors and windows swing open wide outward with strong creaking motion, shutters flapping, hinges stretching, and the whole structure revealing its interior dramatically. \\

\texttt{a306e2ee} & A ship violently breaking apart, hull splintering outward, masts snapping and falling, wood fragments and planks flying in all directions with dramatic destructive wrecking motion. \\

\texttt{ee8ecf65} & A stone angel slowly spreading its wings wide open in a full dramatic extension, feathers fanning outward, arms lifting, and the pose shifting with majestic lifelike motion. \\

\texttt{51b1b31d} & A small castle suddenly igniting as fire breaks out and spreads rapidly across the structure, flames licking upward, smoke billowing, and embers flying with intense burning motion. \\

\texttt{f351569d} & A stone golem slowly bending its massive knees and lowering itself down to sit, stone grinding and cracking, dust falling from joints, with heavy deliberate motion. \\

\texttt{a3d0c28c} & A mech activating and firing its arm-mounted weapon with bright glowing energy blasts, recoil shaking the arm, muzzle flashing, and spent casings ejecting with powerful shooting motion. \\

\texttt{f8a7eafe} & A humanoid mech gripping its rifle firmly and firing rapid glowing shots forward, shoulders bracing against recoil, barrel flashing, with bold aggressive shooting motion. \\

\texttt{dd4c51c1} & A police robot performing energetic dance moves, shifting its weight side to side, swinging its arms rhythmically, head bobbing, and hips swaying with lively exaggerated dance motion. \\

\texttt{be7deb26} & A princess character twirling gracefully with her flowing dress fanning outward wide, butterflies fluttering around her, arms outstretched, hair swirling with elegant spinning dance motion. \\

\texttt{b358d0eb} & Colorful fish swimming actively around a bowl with energetic fin movements, darting and chasing each other, bubbles rising rapidly with lively aquatic motion. \\

\texttt{cd3c309f} & A dragon head on armor breathing a powerful burst of glowing fire forward, flames roaring and flickering intensely, heat shimmer distorting the air, jaws opening wide with fierce dramatic motion. \\

\texttt{fdf979f5} & A potted plant rapidly growing taller with stems shooting upward, leaves and colorful flowers sprouting and expanding quickly, branches spreading outward with fast energetic growing motion. \\

\bottomrule
\end{longtable}

\section{TexVerse test set}
\label{sec:texverse}
We construct a fixed TexVerse test set of 32 held-out assets from the TexVerse-1K dataset. 
The assets are obtained from the official TexVerse Hugging Face repository:
\url{https://huggingface.co/datasets/YiboZhang2001/TexVerse}.
Table~\ref{tab:texverse_test_set} lists the corresponding asset paths in the repository.

\begin{table}[t]
  \centering
  \caption{\textbf{TexVerse test set.}
  List of the 32 held-out TexVerse-1K assets used in our experiments.}
  \label{tab:texverse_test_set}
  \scriptsize
  \setlength{\tabcolsep}{4pt}
  \renewcommand{\arraystretch}{0.7}
  \begin{tabularx}{\linewidth}{cX}
    \toprule
    \textbf{\#} & \textbf{Asset path} \\
    \midrule
    1  & \path{glbs/glbs_1k/000-047/ef95775fcbae453a924071ff4bf3acb1_1024.glb} \\
    2  & \path{glbs/glbs_1k/000-047/ed43cd02f71346ec9313429638db6af7_1024.glb} \\
    3  & \path{glbs/glbs_1k/000-046/eab7714de6d9488cafe6bfba15a1349b_1024.glb} \\
    4  & \path{glbs/glbs_1k/000-074/d9d84b1f6bed433eb69b6ab6d16c3bc5_1024.glb} \\
    5  & \path{glbs/glbs_1k/000-043/dafeb1c2ec924298a2503a8d03150f3a_1024.glb} \\
    6  & \path{glbs/glbs_1k/000-040/c867839965ac44dc975e48f849c03159_1024.glb} \\
    7  & \path{glbs/glbs_1k/000-038/bddeaa8febea4481adb8fb449e372f1e_1024.glb} \\
    8  & \path{glbs/glbs_1k/000-043/d8e3f050f3e4453bb7adf5b0aae229f3_1024.glb} \\
    9  & \path{glbs/glbs_1k/000-085/b9d748ccc31344e2966d3f8176aed656_1024.glb} \\
    10 & \path{glbs/glbs_1k/000-037/bb66a60bf1884d2a930d2bdf2e342fbf_1024.glb} \\
    11 & \path{glbs/glbs_1k/000-030/99d607aef15340a3bfd253b99089d676_1024.glb} \\
    12 & \path{glbs/glbs_1k/000-029/91bf7ee81767422fae8ef8cbc843bf1c_1024.glb} \\
    13 & \path{glbs/glbs_1k/000-027/8a383eb03ef24d4f8df3b68aa121cf9d_1024.glb} \\
    14 & \path{glbs/glbs_1k/000-021/6a1516209ee8411da12acf7b1ec3e98f_1024.glb} \\
    15 & \path{glbs/glbs_1k/000-082/6d4835a9aedc4f369580cc45db413932_1024.glb} \\
    16 & \path{glbs/glbs_1k/000-015/4dd70b77276b4cb195bb01cf269b6c0b_1024.glb} \\
    17 & \path{glbs/glbs_1k/000-016/52f8934bbe1c4da8911c615399f6a8ea_1024.glb} \\
    18 & \path{glbs/glbs_1k/000-013/427bc35ebcf34795be1c1f776cb1307a_1024.glb} \\
    19 & \path{glbs/glbs_1k/000-055/2a4d2840fc904211bab65e2c4a9f52a0_1024.glb} \\
    20 & \path{glbs/glbs_1k/000-080/1eab696ff8934d44b5343747f8cf3eaf_1024.glb} \\
    21 & \path{glbs/glbs_1k/000-003/10a8bbaa7ad648f6a5990b9923e4fa06_1024.glb} \\
    22 & \path{glbs/glbs_1k/000-002/0b80aedfcc8b4f7f99fca567a768d8a5_1024.glb} \\
    23 & \path{glbs/glbs_1k/000-052/099f508befcd4ebe9930502b48978042_1024.glb} \\
    24 & \path{glbs/glbs_1k/000-017/57d76986e57a492a865875b54351ddd7_1024.glb} \\
    25 & \path{glbs/glbs_1k/000-018/5a5bbca8e18744ebacdce8320f014a1e_1024.glb} \\
    26 & \path{glbs/glbs_1k/000-066/8ec8f8a2022d491fa0d74279c4f07304_1024.glb} \\
    27 & \path{glbs/glbs_1k/000-030/9943ce328fff4c298ecdb92d03bdebb5_1024.glb} \\
    28 & \path{glbs/glbs_1k/000-084/92e28edf31074225ad28b652b426b4f4_1024.glb} \\
    29 & \path{glbs/glbs_1k/000-034/aab888ae4c4f4e91ba053c39221ffb05_1024.glb} \\
    30 & \path{glbs/glbs_1k/000-037/ba253b4d995140aca8a968f4a86b2a40_1024.glb} \\
    31 & \path{glbs/glbs_1k/000-042/d6c86c5eeb204bfd985f7bdfb270327d_1024.glb} \\
    32 & \path{glbs/glbs_1k/000-049/f5cf2511b7ce49588ba01565d9afc687_1024.glb} \\
    \bottomrule
  \end{tabularx}
\end{table}

\section{Societal Impacts}
\label{sec:society}
Our work reduces the barrier to creating high-quality 4D generation, which can benefit creators in animation, gaming, and VR/AR applications. 


\end{document}